\documentclass[twocolumn]{article}

\usepackage[T1]{fontenc}
\usepackage{amsmath}
\usepackage{graphicx}
\usepackage{hyperref}
\hypersetup{
    colorlinks,
    linkcolor={red!50!black},
    citecolor={white!25!black},
    urlcolor={blue!70!black}
}

\usepackage{xcolor}
\definecolor{opcol}{HTML}{dd8452}
\definecolor{constcol}{HTML}{c44e52}
\definecolor{kwcol}{HTML}{4c72b0}
\definecolor{typecol}{HTML}{8172b3}

\usepackage{listings}

\lstdefinelanguage{pseudolang}{
    keywords={def,for,parfor,in,yield,if,else,while,break,atomic},
    morecomment=[l]{//}
}
\lstdefinelanguage{opencl}{
    keywords={channel,\_\_attribute\_\_,void,for,break,if,else,while,true,false,\_\_kernel,\_\_global},
    morecomment=[l]{//}
}
\lstdefinestyle{pseudocode}{
  language=pseudolang,
  keywordstyle=\color{kwcol}\bfseries,
  commentstyle=\color[HTML]{55a868}\itshape,
  basicstyle=\ttfamily,
  numberstyle=\scriptsize\ttfamily,
  columns=fullflexible,
  keepspaces=true,
  numbers=left,
  numbersep=6pt,
  xleftmargin=1.5em,
  morecomment=[l]{\{},
  emph={range,True,False,write,read},
  emphstyle={\color{typecol}},
  literate={+=}{{{\color{opcol}+=}}}2 {+}{{{\color{opcol}+}}}1 {-}{{{\color{opcol}-}}}1 {\%}{{{\color{opcol}\%}}}1 {<}{{{\color{opcol}<}}}1 {/}{{{\color{opcol}/}}}1 {*}{{{\color{opcol}*}}}1 {=}{{{\color{opcol}=}}}1 {D\_MAX}{{{\color{constcol}D\_MAX}}}1 {H}{{{\color{constcol}H}}}1 {N\_CLS}{{{\color{constcol}N\_CLS}}}1 {DONE}{{{\color{constcol}DONE}}}1
}

\lstdefinestyle{opencl}{
  language=opencl,
  basicstyle=\ttfamily,
  numberstyle=\scriptsize\ttfamily,
  columns=fullflexible,
  keepspaces=true,
  numbers=left,
  numbersep=6pt,
  xleftmargin=1.5em,
  keywordstyle=\color{kwcol}\bfseries,
  emph={int,char,double,float,unsigned,bool,uint},
  emphstyle={\color{typecol}},
  literate={+=}{{{\color{opcol}+=}}}2 {+}{{{\color{opcol}+}}}1 {-}{{{\color{opcol}-}}}1 {\%}{{{\color{opcol}\%}}}1 {<}{{{\color{opcol}<}}}1 {/}{{{\color{opcol}/}}}1 {*}{{{\color{opcol}*}}}1 {=}{{{\color{opcol}=}}}1 {D\_MAX}{{{\color{constcol}D\_MAX}}}1 {H}{{{\color{constcol}H}}}1 {N\_CLS}{{{\color{constcol}N\_CLS}}}1 {DONE}{{{\color{constcol}DONE}}}1 {!}{{{\color{constcol}!}}}1
}

\usepackage{natbib}
\usepackage{booktabs}
\usepackage[noabbrev]{cleveref}
\usepackage{subcaption}

\author{Björn A. Lindqvist and Artur Podobas\\ KTH Royal Institute of Technology \\ Stockholm, Sweden}
\title{Fast Algorithms for Spiking Neural Network Simulation with FPGAs}

\begin{document}

%% \onecolumn
%% \firstpage{1}
%% \title[SNNs on FPGAs]{Design, Implementation, and Evaluation of a Highly-Connected Cortical Microcircuit Model on a Field-Programmable Gate Array}
%% \author[\firstAuthorLast ]{\Authors}
%% \address{}
%% \correspondance{}
%% \extraAuth{}

\maketitle

\begin{abstract}
  \noindent Using OpenCL-based high-level synthesis, we create a
  number of spiking neural network (SNN) simulators for the
  Potjans-Diesmann cortical microcircuit for a high-end
  Field-Programmable Gate Array (FPGA). Our best simulators simulate
  the circuit 25\% faster than real-time, require less than 21 nJ per
  synaptic event, and are bottle-necked by the device's on-chip
  memory. Speed-wise they compare favorably to the state-of-the-art
  GPU-based simulators and their energy usage is lower than any other
  published result. This result is the first for simulating the
  circuit on a single hardware accelerator. We also extensively
  analyze the techniques and algorithms we implement our
  simulators with, many of which can be realized on other types of
  hardware. Thus, this article is of interest to any researcher or
  practitioner interested in efficient SNN simulation, whether they
  target FPGAs or not.
\end{abstract}

\section{Introduction}

\cite{natkey1} outline what they see as the four biggest challenges
for human brain simulation; scale, complexity, speed, and
integration. Scale refers to the enourmous size of the brain --
billions of neurons and trillions synapses -- which is difficult to
simulate at acceptable speeds even for the fastest
supercomputers. There are two reasons for this difficulty. First, the
conceptual mismatch between how brains and computers operate; the
former can be viewed as collections of billions of processing nodes,
communicating with sparse events called spikes \citep{ghosh2009third},
while the latter is composed of imperative programs, vulnerable to the
von Neumann bottleneck~\citep{efnusheva2017survey}, and executed on
general-purpose devices (GPUs or GPUs). Power \textit{in-efficiency}
is the second reason. General-purpose devices are jacks of all trades
and can execute a wide variety of workloads, ranging from word
processing to scientific simulations. Their generality comes at a
high price and executing a single instruction consumes up to three
orders of magnitude more energy than the computation
itself~\citep{jouppi2018motivation}. This causes brain simulations to
consume far more energy than the roughly 20 Watts a human brain
draws~\citep{versace2010brain}. Finally, the ~2004 end of Dennard's
scaling~\citep{bohr200730}, and the impending termination of Moore's
law~\citep{theis2017end}, forces researchers to reconsider how to
compute efficiently in a post-Moore future. Among the more promising
options in a post-Moore world is the use of reconfigurable systems
such as Field-Programmable Gate Arrays (FPGAs)~\citep{kuon2008fpga}.

FPGAs, along with their siblings, Coarse-Grained Reconfigurable
Systems, CGRAs~\citep{podobas2020survey}, belong to the reconfigurable
family of computing devices. Unlike traditional general-purpose
processors (CPUs and GPUs), their underlying compute fabric is
composed of a large number of reconfigurable blocks of different
types. The most common blocks are look-up tables (LUTs, often several
hundred thousands), digital signal processing (DSPs -- capable of tens
of TFLOP/s), or on-chip random-access memory (BRAM -- tens-to-hundreds
of MB)~\citep{langhammer2021stratix,murphy2017xilinx}. These resources
allow designers to create custom hardware that sacrifice generality
for better performance and lower energy consumption than
general-purpose devices, as well as transcending the latters'
limitations (i.e., the mentioned von Neumann-bottleneck). Importantly,
these devices can, and often already are
~\citep{sano2023essper,meyer2023multi,boku2022cygnus}, live
side-by-side in HPC nodes and can be reconfigured before
runtime~\citep{vipin2018fpga}: when the user is running a brain
simulation, the FPGA will be configured as an efficient brain
simulator, while for a different application a different accelerator
will be used. In short, FPGA facilitates the use of special hardware
accelerators during application runtime but is general enough so that
different accelerators can be configured between applications. This
makes them an attractive choice for neuroscience simulation since they
can be reconfigured for different neuron, synapse, and axon models,
which dedicated neuromorphic ASIC hardware, whose silicon is
immutable, cannot be.

Historically, FPGAs have been designed using low-level hardware
description languages (HDLs) such as VHDL and
Verilog~\citep{perry2002vhdl}. These languages have a steep learning
curve and require specialist knowledge to be comfortable
with. However, with the increased maturity of High-Level Synthesis
(HLS) tools in the last two decades~\citep{nane2015survey}, there has
been a resurgence of interest in using FPGAs for HPC. HLS allow
designers to describe hardware in relatively high-level
languages such as C and C++ whose learning curves are shallower
~\citep{podobas2017evaluating,zohouri2016evaluating}. For example,
FPGAs have been used to accelerate Computational Fluid
Dynamics~\citep{karp2021high,faj2023scalable}, Quantum circuit
simulations~\citep{podobas2023q2logic,aminian2008fpga}, Molecular
Dynamics~\citep{sanaullah2018unlocking}, N-Body
systems~\citep{del2018scalable,menzel2021strong,huthmann2019scaling},
and much more, demonstrating advantages over alternative
solutions.

In this work, we use of HLS to design simulators for the
Potjans-Diesmann cortical microcircuit~\citep{potjans2014}. While
there is ample prior work on FPGA-based neuromorphic systems (see
\cref{sec:sota} for related work), \textit{our system is (to the best
  of our knowledge) the most energy-efficient simulator of the
  Potjans-Diesmann circuit in existence (25 nJ/event), while reaching
  a faster-than-realtime ($\approx$ 1.2x) simulation speed on a single
  FPGA}. We use the Intel's OpenCL SDK for FPGA HLS
toolchain~\citep{czajkowski2012opencl} to design our simulators, but
our designs are modular enough to easily be ported to other HLS-based
systems (e.g., Vivado~\citep{o2014xilinx} or OneAPI), and other
FPGAs. Our contributions are:

\begin{itemize}
  \item The first simulators of the previously mentioned circuit on a
    single FPGA, running faster than real-time.
  \item The most energy-efficient simulators for the circuit when
    measured by energy per synaptic event.
  \item The presentation and analysis of the algorithms, thought
    processess, trade-offs, and lessons learned, while designing these
    simulators.
  \item An empirically motivated analysis on what hardware features
    are required to simulate the circuit even faster than what we are
    capable of.
\end{itemize}

The rest of this article is structured as follows. In \cref{sec:mats},
we discuss SNNs in general and the microcircuit in particular. We
explain how FPGAs work and we briefly introduce the HLS design
methodology. In \cref{sec:ssn-sim} we discuss SNN simulation and
present the algorithms and ideas underlying our simulators. The
utility of many of the ideas have already been demonstrated
in other parts of Computer Science, but not in connection with SNN
simulation. Hence, we believe they deserve a thorough treatment. We
evaluate many different variants and parametrizations of our
simulators in \cref{sec:results}. Finally, in \cref{sec:disc} we put
our results in perspective and compare them with the state-of-the-art.

\section{Material and Methods}
\label{sec:mats}

We begin with an overview of spiking neural networks (SNNs) before
discussing the Potjans-Diesmann cortical microcircuit, an SNN for
simulating a small part (microcircuit) of the mammalian brain. In
\cref{sec:fpga} we explain how FPGAs works and what makes them
different from from conventional hardware. In sections
\cref{sec:hls,sec:ocl,sec:ocl-fpga} we introduce HLS and how we use
OpenCL for HLS.

\subsection{Spiking Neural Networks}

An SNN is an artificial neural network (ANN) that transfers signals in
time-dependent bursts, i.e. spikes. Unlike other ANNs, SNNs are
designed with biological plausibility in mind, making them useful for
neuroscience. SNNs are usually modelled as directed (multi-)graphs,
where vertices represent neurons and edges synaptic connections
between neurons. Neurons have a membrane potential that varies over
time. When the potential exceeds a threshold the neuron discharges --
spikes -- and sends current via its synapses to its neighbours which
they receive after a synapse-specific delay. The amount of current as
well as the transfer time is synapse-specific \citep{han2020}. The
neuron's statefulness and the non-differentiable, discontinuous signal
transfer function are two fundamental aspects distinguishing SNNs
from other ANNs. While these basic operating principles are enough to
describe most SNNs, SNNs vary in neuron model and other parameters. In
this work, we use the basic leaky integrate-and-fire (LIF) neuron
model, defined by
\begin{equation}
  RC\frac{\mathrm{d}u}{\mathrm{d}t} = u_{\mathrm{rest}} - u(t) + RI(t).
  \label{eqn:lif}
\end{equation}
The equation describes the membrane potential over time. The
variables $R$ and $C$ are the resistance and capacitance of the
membrane, $u(t)$ its potential at time $t$, $I(t)$ the amount of
current it receives at time $t$ from its neighbours, and
$u_{\mathrm{rest}}$ its resting potential. With $\tau_m = RC$, and
$u(0) = u_{\mathrm{rest}} = 0$ the solution to the equation is
\begin{equation}
  u(t) = RI(t) - RI(t)e^{-\frac{t}{\tau_m}}.
\end{equation}
Solving the equation using forward Euler produces a recurrent,
discrete representation of the neuron's potential over time. With
$\Delta t \to 0$,
\begin{align}
  \tau_m\frac{u(t + \Delta t) - u(t)}{\Delta t} = -u(t) + RI(t)\\
  \implies u(t + \Delta t) = u(t) + \frac{\Delta t}{\tau_m}\left(-u(t) + RI(t)\right)\\
  \implies u(t + \Delta t) = \left(1 - \frac{\Delta t}{\tau_m}\right)u(t) + \frac{\Delta tR}{\tau_m}I(t).
\end{align}
The solution forms the basis of step-wise simulation of LIF
neurons.

After a neuron spikes it enters its refractory state. Its potential
becomes fixed at $u_{\mathrm{reset}}$ and it ceases to respond to
stimuli for a duration controlled by the $\tau_{\mathrm{ref}}$
parameter. Usually, the refractory period is in the order of
milliseconds and for simplicity one sets $u_{\mathrm{reset}} =
u_{\mathrm{rest}} = 0$. With $r(t)$ denoting how long
the neuron will say refractory at time $t$, $u_{\mathrm{thr}}$ the
neuron's spiking threshold, and $\Delta t$ arbitrarily set to one, we
can incorporate refractoriness into the LIF model:
\begin{align}
  r(t + 1) &= \begin{cases}
    \tau_{\mathrm{ref}}&\mathrm{if}\,u(t + 1) \geq u_{\mathrm{thr}}\\
    r(t) - 1&\mathrm{elseif}\,r(t) > 0\\
    0&\mathrm{otherwise}\\
  \end{cases}\\
  u(t + 1) &= \begin{cases}
    \left(1 - \frac{1}{\tau_m}\right)u(t) + \frac{R}{\tau_m}I(t)&\mathrm{if}\quad r(t)=0\\
    0&\mathrm{otherwise}
  \end{cases}
  \label{eqn:sync-u}
\end{align}

\subsubsection{Network topology}
\label{sec:topo}
\begin{figure}
  \center
  \includegraphics[width=\linewidth]{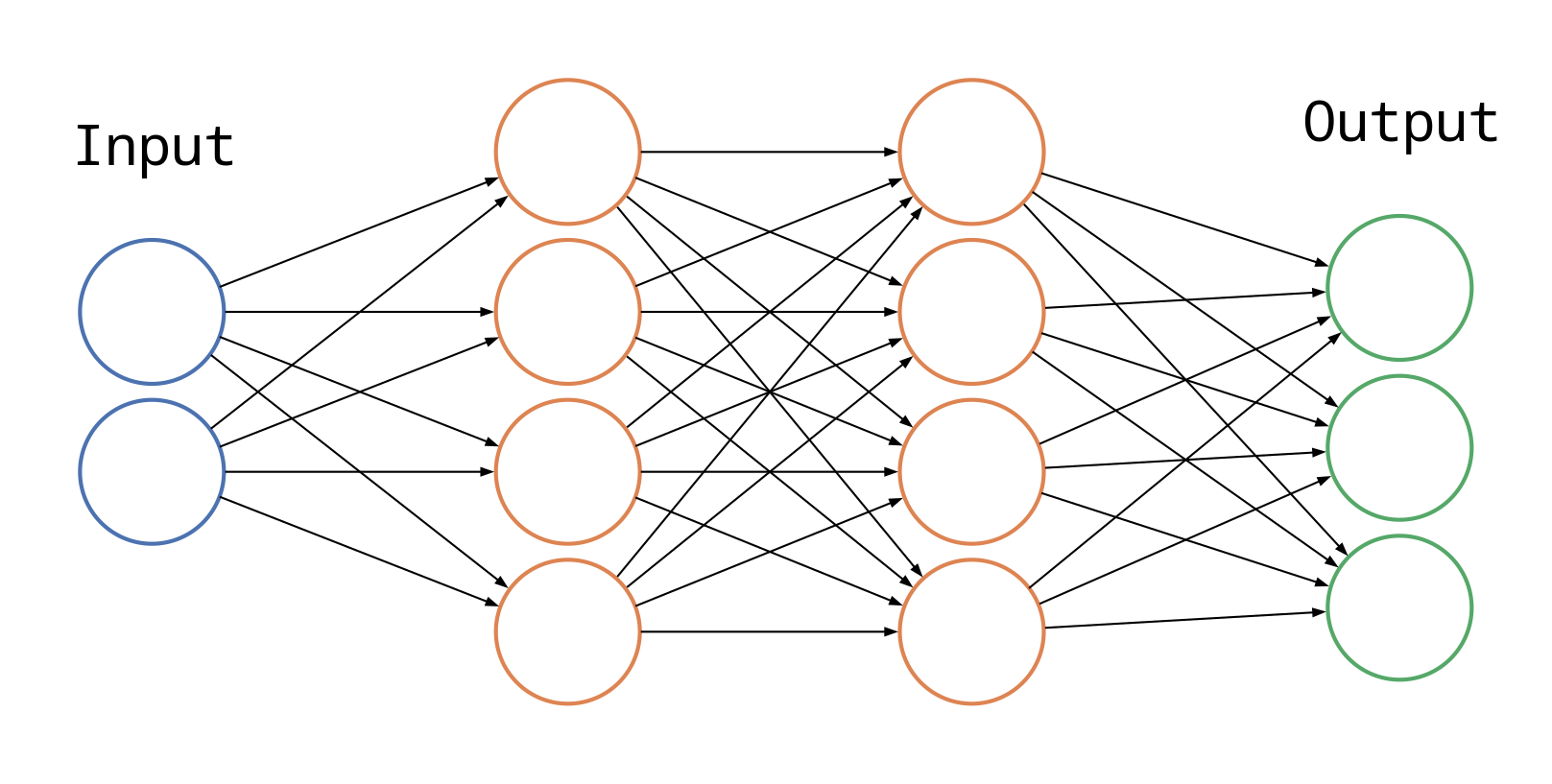}
  \caption{A fully-connected neural network with two hidden layers,
    two input neurons, and three output neurons}
  \label{fig:fcn}
\end{figure}

A major difference between conventional ANNs and SNNs is that the
former often are layered; all neurons in one layer only receives
inputs from neurons in the previous layer and only sends outputs to
neurons in the following layer. If all neurons in a layer send output
only to all neurons in the following layer the layer is said to be
fully-connected and its signal-transfer can be represented as a
matrix-vector multiplication. SNNs can similarily be layered and it
has been found to be a good approach for classification
\citep{zheng2021}. But it is not suitable for simulation as the
neurons in real brains are not organized into fully-connected
layers. Instead, their topology is ``chaotic'' and full of recurrent
connections, self-loops (autapses), and multiple edges
(multapses). This has far-reaching consequences for what data
structures are appropriate for SNNs. An adjacency matrix, for example,
is not enough to represent their topological richness.

\subsection{Potjans-Diesmann's Microcircuit}
\begin{table}
  \centering
  \footnotesize
  \begin{tabular}{lll}
    \toprule
    \textbf{Name} & \textbf{Value} & \textbf{Description (Unit)}\\
    \midrule
    $\Delta t$ & 0.1 & Time step duration (ms)\\
    $C_m$ & 250 & Membrane capacity (pF)\\
    $\tau_m$ & 10 & Membrane time constant (ms)\\
    $\tau_{\mathrm{ref}}$ & 2 & Refractory period (ms)\\
    $\tau_{\mathrm{syn}}$ & 0.5 & Postsyn. current time constant (ms)\\
    $u_{\mathrm{rest}}$ & -65 & Resting and reset potential (mV)\\
    $u_{\mathrm{thr}}$ & -50 & Spiking threshold (mV)\\
    $v_{\mathrm{th}}$ & 8 & Thl. neurons' mean spiking rate (Hz)\\
    $\omega_{\mathrm{ext}}$ & 0.15 & Thl. spikes amplitude (mV)\\
    \bottomrule
  \end{tabular}
  \caption{Microcircuit's general and simulation parameters}
  \label{tab:params}
\end{table}

\begin{table*}[t]
  \centering
  \begin{tabular}{rlrrlll}
    \toprule
    $i$ & \textbf{Pop.} & $N_i$ & $K$ & $u_\mathrm{init}$ & $\omega_i$ & $\delta_i$ \\
    \midrule
    1 & L23/exh & 20 683 & 1 600 & $\mathcal{N}(-68.28, 5.36)$ & $\mathcal{N}(0.15, 0.015)$ & $\mathcal{N}(1.5, 0.75)$\\
    2 & L23/inh &  5 834 & 1 500 & $\mathcal{N}(-63.16, 4.57)$ & $\mathcal{N}(-0.6, 0.06)$ & $\mathcal{N}(0.75, 0.325)$\\
    3 &  L4/exh & 21 915 & 2 100 & $\mathcal{N}(-63.33, 4.74)$ & $\mathcal{N}(0.15, 0.015)$ & $\mathcal{N}(1.5, 0.75)$\\
    4 &  L4/inh &  5 479 & 1 900 & $\mathcal{N}(-63.45, 4.94)$ & $\mathcal{N}(-0.6, 0.06)$ & $\mathcal{N}(0.75, 0.325)$\\
    5 &  L5/exh &  4 850 & 2 000 & $\mathcal{N}(-63.11, 4.94)$ & $\mathcal{N}(0.15, 0.015)$ & $\mathcal{N}(1.5, 0.75)$\\
    6 &  L5/inh &  1 065 & 1 900 & $\mathcal{N}(-61.66, 4.55)$ & $\mathcal{N}(-0.6, 0.06)$ & $\mathcal{N}(0.75, 0.325)$\\
    7 &  L6/exh & 14 395 & 2 900 & $\mathcal{N}(-66.72, 5.46)$ & $\mathcal{N}(0.15, 0.015)$ & $\mathcal{N}(1.5, 0.75)$\\
    8 &  L6/inh &  2 948 & 2 100 & $\mathcal{N}(-61.43, 4.48)$ & $\mathcal{N}(-0.6, 0.06)$ & $\mathcal{N}(0.75, 0.325)$\\
    \bottomrule
  \end{tabular}
  \caption{Population-specific parameters. The first four columns
    denote the index, $i$, the name, the size, $N_i$, and the number
    of thalamic connections, $K_i$, of the eight populations. The last
    three columns denote the Gaussians from which the neurons' initial
    potential (mV), the neurons' synapses amplitudes (mV), and delays
    (ms) of excitatory postsynaptic potential are sampled
    from. However, synapse amplitudes from population L23/exh to L3/exh are
    sampled from $\mathcal{N}(0.3, 0.03)$ and not $\mathcal{N}(0.15,
    0.015)$.}
  \label{tab:params2}
\end{table*}

\begin{table*}[t]
  \centering
  \begin{tabular}{lrrrrrrrr}
    \toprule
    & L23/exc & L23/inh & L4/exc & L4/inh & L5/exc & L5/inh & L6/exc & L6/inh\\
    \midrule
    L23/exc & 0.1009 & 0.1689 & 0.0437 & 0.0818 & 0.0323 & 0.0000 & 0.0076 & 0.0000\\
    L23/inh & 0.1346 & 0.1371 & 0.0316 & 0.0515 & 0.0755 & 0.0000 & 0.0042 & 0.0000\\
    L4/exc & 0.0077 & 0.0059 & 0.0497 & 0.1350 & 0.0067 & 0.0003 & 0.0453 & 0.0000\\
    L4/inh & 0.0691 & 0.0029 & 0.0794 & 0.1597 & 0.0033 & 0.0000 & 0.1057 & 0.0000\\
    L5/exc & 0.1004 & 0.0622 & 0.0505 & 0.0057 & 0.0831 & 0.3726 & 0.0204 & 0.0000\\
    L5/inh & 0.0548 & 0.0269 & 0.0257 & 0.0022 & 0.0600 & 0.3158 & 0.0086 & 0.0000\\
    L6/exc & 0.0156 & 0.0066 & 0.0211 & 0.0166 & 0.0572 & 0.0197 & 0.0396 & 0.2252\\
    L6/inh & 0.0364 & 0.0010 & 0.0034 & 0.0005 & 0.0277 & 0.0080 & 0.0658 & 0.1443\\
    \bottomrule
  \end{tabular}
  \caption{Probability that a random neuron in the population specified by the rows is connected to a random neuron in the population specified by the columns.}
  \label{tab:params3}
\end{table*}

In 2014 \cite{potjans2014} compiled the results of a dozen empirical
studies to create a full-scale spiking neural network microcircuit of
one cortical column in the mammalian early sensory cortex. The
microcircuit covers 1 $\mathrm{mm}^2$ of the cerebral cortex and
consists of 77,169 LIF neurons grouped into four neocortical
layers;\footnote{These layers are \textit{not} analoguous to layers in
conventional ANNs.} L23, L4, L5, and L6.\footnote{Layer L2 and L3
merged and L1 omitted.}  Each layer is subdivided into one excitatory
population that increases neural activity and one inhibitory
population that decreases it. Around 300 million synapses connect the
neurons.

The microcircuit is a balanced random network so that neural activity
is balanced by excitatory neurons that inreases activity and
inhibitory ones that dampen it \citep{brunel2000}. Connectivity and
features are sampled from parametric probability distributions, rather
than set explicitly. \Cref{tab:params2} and \cref{tab:params3} specify
these distributions' parameters. For example, the initial potential of
every inhibitory neuron in L23 is set by sampling a Gaussian with mean
-63.16 mV and standard deviation 4.57 mV and the expected number of
synapses from population L23/inh to population L5/exc is $5,834 \cdot
4,850 \cdot 0.0755 \approx 2$ million. Synapses are sampled with
replacement -- multiple synapses can connect the same neuron
pair. While the neurons are arranged in in terms of neocortical
layers, the layers' connection probabilities show that the network's
\textit{topology} is not layered; neurons in most populations can
connect to neurons in any of the other populations.

In addition to the synapses within the cortical column, the circuit
receives spikes from external neurons -- thalamic input. Column $K$ in
\cref{tab:params2} specifies the number of thalamic neurons a given
population's neurons receive spikes from, and parameter
$v_{\mathrm{th}}$ in \cref{tab:params} how frequently thalamic neurons
spike.\footnote{\cite{potjans2014} set the parameter to 15 Hz, but we
use the NEST model as a baseline, where it is set to 8 Hz.}
The expected number of thalamic spikes received per second by all
neurons in a population is $v_\mathrm{th}K$. For example, neurons in
population L23/exc receive about $8\cdot1,600 = 12,800$ thalamic
spikes per second. The amplitude of all thalamic synapses is fixed at
$\omega_\mathrm{exh}=0.15$ mV. As thalamic spikes can be
computationally expensive to simulate, \cite{potjans2014} suggest
approximating them with constant direct current injected at a rate of
$v_{\mathrm{th}}K\omega_\mathrm{exh}\tau_{\mathrm{syn}}$ mV per
second. In our simulator we model thalamic spikes, however.

The synaptic parameters are also scaled. The synapse amplitude by
$w_f$ which is a function of the membrane time constant, $\tau_m$,
membrane capacity, $C_m$, and the postsynaptic time constant,
$\tau_{\mathrm{syn}}$, that maps postsynaptic potential to
postsynaptic current:\footnote{See \cite{hanuschkin2010} for the derivation.}
\begin{align}
  d &= \tau_{\mathrm{syn}} - \tau_m\\
  p &= \tau_{\mathrm{syn}}\tau_m\\
  q &= \tau_m / \tau_{\mathrm{syn}}\\
  w_f &= \frac{C_md}{p(q^{\tau_m/d} - q^{\tau_{\mathrm{syn}}/d})} \approx 585
\end{align}
The constants $p_{22}$ and $p_{11}$ define the membranes' and
presynaptic currents' decay rate:
\begin{align}
  p_{11} &= \exp(-\Delta t/\tau_{\mathrm{syn}}) \approx 0.82\\
  p_{22} &= \exp(-\Delta t/\tau_m) \approx 0.99
  \label{eqn:p11p22}
\end{align}
The injection of the presynaptic current is scaled by $p_{21}$:
\begin{align}
  \beta &= \tau_{\mathrm{syn}}\tau_m / (\tau_m - \tau_{\mathrm{syn}})\\
  \gamma &= \beta / C_m\\
  p_{21} &= p_{11}\gamma(\exp(\Delta t/\beta) - 1) \approx 0.00036
  \label{eqn:p21}
\end{align}
The subscripts match those found in the source code for the NEST
simulator \citep{plesser2015} and have no deeper meaning
here.\footnote{\url{https://github.com/nest/nest-simulator/blob/aa9907a5d5a7916eeeeca00c2e6584202702eb2a/models/iaf_psc_exp.cpp}} Taken together, this gives us the following discrete recurrences for
the step-wise update of membrane potential, $u_t$:
\begin{equation}
  u_{t + 1} = \begin{cases}
    u_{\mathrm{reset}} & \text{if}\quad r_t > 0\\
    p_{22}u_t + I_tp_{21} & \text{otherwise}\\
  \end{cases},
  \label{eqn:mem}
\end{equation}
presynaptic current, $I_t$:
\begin{equation}
  I_{t + 1} = p_{11}I_t + T_tw_f\omega_\mathrm{exh},
  \label{eqn:psc}
\end{equation}
and refractoriness, $r_t$:
\begin{equation}
  r_{t + 1} = \begin{cases}
    \tau_{\mathrm{ref}} & \text{if}\quad u_{t + 1} \geq u_{\mathrm{thr}}\\
    r_t - 1 & \text{elseif}\quad r_t > 0\\
    0 & \text{otherwise}
  \end{cases}.
  \label{eqn:ref}
\end{equation}
The variable $T_t$ denotes the number of thalamic spikes received by
the presynapse at time $t$ and is modelled as a Poisson distributed
random variable with mean $v_{\mathrm{th}}K\Delta t$.

Having presented the theoretical foundations for LIF SNNs and the
specifics of the Potjans-Diesmann microcircuit, we now present the
technologies implement our simulator with. We return to SNN simulation
in \cref{sec:ssn-sim} where we both delve deep into simulation methods
and present our simulators.

\subsection{Field-Programmable Gate Arrays}
\label{sec:fpga}

\noindent An FPGA is a type of reprogrammable integrated
circuit. First marketed by Altera and Xilinx in the 1980's, FPGAs have
found uses in many niches of the electronics industry; in avionics, in
telecommunications, and in VLSI design because of their unique
blend of performance, flexibility, and development costs
\citep{trimberger2015}. FPGAs are not as performant as
Application-Specific Integrated Circuits (ASICs), but ASICs are
extremely expensive to develop which makes them cost-prohibitive
unless the number of units produced runs in the tens of
millions. Furthermore, ASICs are not reprogrammable and thus suitable
only for the specific tasks they were developed for. Central
Processing Units (CPUs), on the other hand, are flexible and
inexpensive, but have low performance. Graphics
Processing Units (GPUs) sit between CPUs and ASICs on the
flexibility-performance spectrum. While GPUs can run any computation,
they generally only excel at highly regular, numerically intensive
computations. In particular, they do not handle divergent control flow
well.

FPGAs trade performance for flexibility in a different manner than
CPUs, GPUs, and ASICs. They consist of configurable blocks organized
in a grid-like fabric and each block can be configured to compute a
small and specific function. Like the logical and of two one-bit
signals or comparing two eight-bit numbers. The number of configurable
blocks varies enormously from one FPGA to another; high-end FPGAs can
contain hundreds of thousands or even millions of blocks. For
performance-sensitive workloads, FPGAs' advantage is the lack
instruction processing overhead. CPUs and GPUs load programs from
memory and decode and process their instructions one after the other,
while FPGAs merely pass data through a pre-configured, fixed-function
circuit, similar to how ASICs operate.\footnote{It is of course true
that superscalar and pipelined processors can handle many instructions
in parallel. But the point stands; the overhead caused by the need to
decode and dispatch instructions is large.} However, the
reprogrammability of FPGAs comes at a great cost and their raw
performance cannot rival that of ASICs or GPUs unless algorithms are
specifically designed for them. They also run at lower clock
frequencies than comparable CPUs from the same generation. For
example, our simulators run at around 600 MHz,\footnote{However,
\cite{langhammer2023} ran a softcore microprocessor at over 770 MHz on
the same FPGA which is close to the theoretical limit.} while Intel
Core i9 processors operate at over 3 GHz. To stay competitive, FPGA
implementations need to carry out much more work per clock cycle than
equivalent CPU implementations. These potential drawbacks aside,
researchers have deployed FPGAs for performance sensitive workloads,
with promising results \citep{nguyen2022}.

Bitstreams are files that configure FPGA blocks and specialized tools
write them to the FPGA's fabric. Bitstreams are to FPGAs what machine
code are to microprocessors. Tools that take descriptions of the
digital circuit the FPGA should implement and produce bitstreams are
called \textit{synthesizers}. The descriptions are usually written in
hardware description languages (HDLs), but many tools support
higher-level languages as well.

\begin{figure}
  \center
  \includegraphics[width=\linewidth]{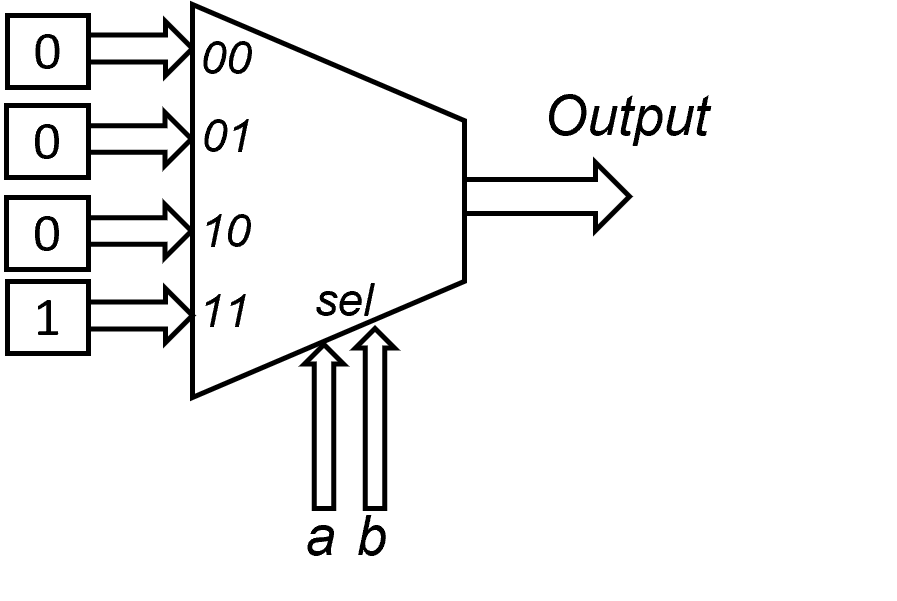}
  \caption{A LUT for computing any two-variable boolean function
    implemented as one 4:1 multiplexer and four memory bits. The bit
    values determines which function the LUT computes.}
  \label{fig:lut}
\end{figure}

A configurable block is a logic block that functions as a small
combinational circuit that computes the boolean function it was
configured for. Depending on configuration, the same logic block can
serve as an and-gate, or-gate, xor-gate, etc. Memory embedded adjacent
to the block stores its configuration and the block's output is
retrieved from this memory. Because the blocks look up their results
from memory they are called look-up-tables (LUTs). \Cref{fig:lut}
shows a LUT for a two-input and-gate built with a multiplexer and
four memory bits. Flip-flops (FFs) are the FPGA's basic memory blocks
and work as small and fast RAMs. Each FF stores one bit, but multiple
FFs can be combined into registers to store larger datums.

In addition to configuring the FPGA's logic and memory blocks,
bitstreams define how the blocks should be connected through the
FPGA's interconnect network. This is called \textit{routing} and is
one of the synthesizer's most important tasks. Routing is critical to
the design as the interconnect occupies most of the FPGA's fabric
\citep{betz1999}. Good routing should minimize the total length of the
interconnect, the number of blocks required, and the length of the
longest path connecting two blocks as the operating frequency of the
design is bounded by this length. Routing is a challenging problem in
computer science and finding high-quality routing of large designs is
both difficult and time-consuming.

Due to their reconfigurability, digital logic implemented in LUTs is
slower and requires more components than equivalent logic implemented
in non-reconfigurable ASIC gates. Memories built using FFs have lots
of overhead because FFs only store one bit. Therefore, modern FPGAs
come with non-reconfigurable blocks for arithmetic and storage. One
can view these blocks as small ASICs embedded in the FPGA that the
designer connects via the configurable blocks. For example, our Agilex
7 FPGA have five different types of blocks; 487 200 Adaptive Logic
Modules (ALMs) which functions as LUTs, 1 948 800 flip-flops, 7 110
M20K RAM blocks, two 18.432 Mb eSRAM blocks, and 4 510 DSPs. We
describe these blocks in the following sections.

\subsubsection{Adaptive Logic Module}
\begin{figure*}
  \includegraphics[width=\linewidth]{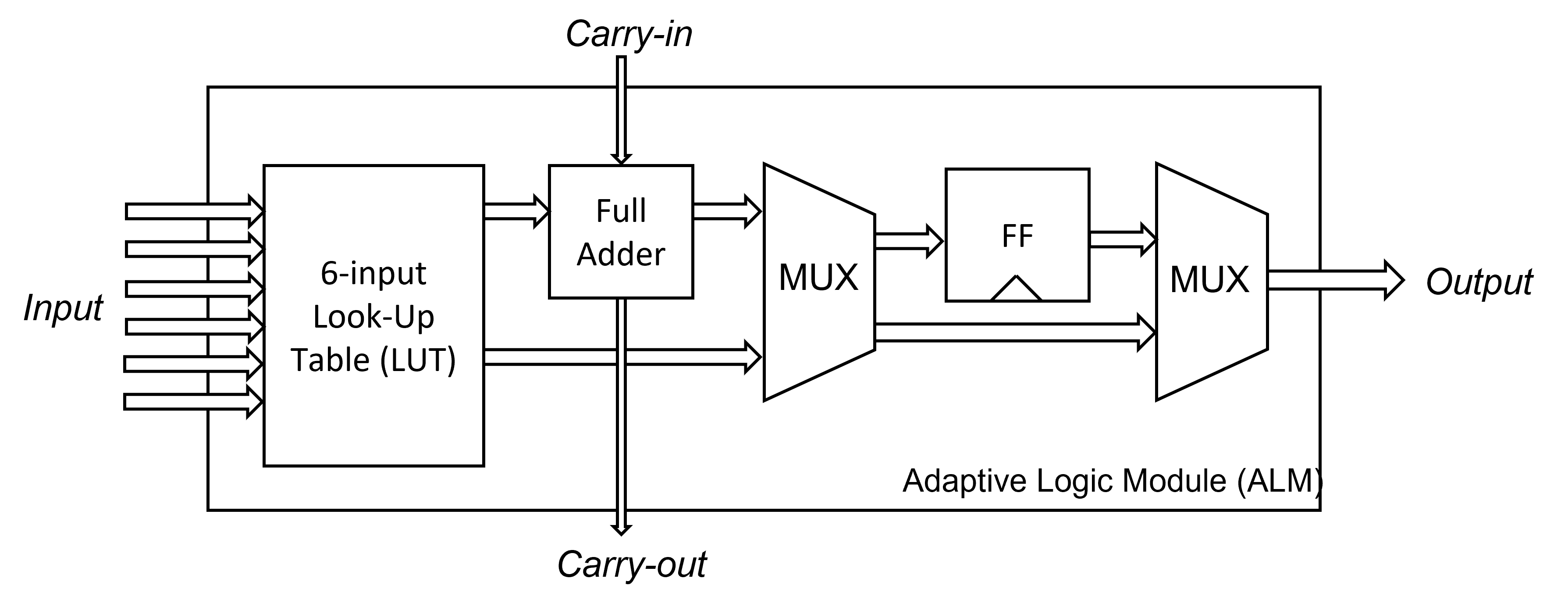}
  \caption{Simplified schematic of the Agilex 7 ALM. The multiplexers'
    (trapezoids) control signals (not shown) and the contents of
    the LUT configures the ALM. The ALM can serve as -- among
    other things -- a four-bit adder, a four-bit memory, or as
    combinational logic of six inputs depending on configuration.}
  \label{fig:alm}
\end{figure*}

The \textit{Adaptive Logic Module} (ALM) in \cref{fig:alm} is the
basic building block of Intel's family of FPGAs. Consisting of one
fractured eight-input-LUT, four FFs, and two full-adders, it is a
versatile multi-purpose block and can -- depending on its
configuration -- compute two four-input boolean functions, one
six-input boolean function, or perform four-bit addition with
carry. It can also serve as a four-bit memory.

\subsubsection{Digital Signal Processing}
\noindent The \textit{Digital Signal Processing} (DSP) block contains
functions for multiplication, addition, subtraction, and
accumulation. The block functions as the FPGA's arithmetic logic unit
(ALU). Agilex 7 has two types of DPSs; one for integer arithmetic and
one for IEEE 754 floating-point arithmetic in single- and
half-precision modes. Pipeline registers organized into three stages
are contained within the DSP. Data can be routed through one or more
of the stages or bypassed completely to achieve a given latency. This
can be useful if one operand is available multiple cycles before the
other.

\subsubsection{Memory Hierarchy}

In processor-based hardware, registers, caches, and main memory is
organized into a fixed hierarchy. Applications must be structured around
the memory hierarchy to optimally use it. Typically,
data transfer between the hierarchy's levels is implicit
and not directly under the programmer's control. On an FPGA the
designer constructs the memory hierarchy from the available
memory blocks and has fine-grained control over exactly where data is
stored. This allows the memory system to be
tailored to the application rather than the other way around.

The FPGA's memory is \textit{on-chip} if it is embedded in the FPGA
fabric itself and \textit{off-chip} otherwise. On-chip memory is much
faster and smaller than off-chip memory. The Agilex 7 has three types
of on-chip memory; Memory Logic Array Blocks (MLABs), M20Ks, and
embedded SRAM (eSRAM). Scalars and shallow FIFOs are often stored in
MLABs, small arrays and caches in M20Ks, and larger buffers in
eSRAM. MLABs are not memory blocks per se, but instead a technique for
using ALMs as memory. The Agilex 7 can combine ten unused ALMs into
one 640-bit register. As many FPGA designs do not use all logic
blocks, repurposing them can be very useful. MLABs have very low
laencies since they are close to the logic that use them. Xilinx FPGAs
implement the same concept under the name \textit{distributed RAM} or
LUTRAM. On-chip M20Ks are also known as block RAM (BRAM) and are much
larger than MLABs. The Agilex 7 contains around seven thousand M20K
each of which can hold 20 kbit of data, as their name suggets. In
total, about 139 Mbit. The board also has two 36 Mbit embedded SRAM
blocks. These have high bandwidth and high random transaction rates
and compliment the other on-chip memory blocks. Unfortunately, the
Intel OpenCL FPGA compiler cannot infer eSRAM so they are unused in
this work. On the Agilex 7 four 8 GB DRAM sticks constitute the
board's off-chip memory. Accessing off-chip memory takes much longer
than accessing on-chip memory, with access times measured in the
hundreds of cycles.

In addition to controlling the types of memory used, the designer can
also configure individual memory blocks. Multipumping, for example,
can sometimes double memory throughput at the expense of significantly
lowering the design's operating frequency. Banking can increase
expected memory access concurrency, but also increase stalls due to
conflicts so the technique is best reserved for evenly distributed
data.

\subsection{High-Level Synthesis}
\label{sec:hls}

Designers traditionally design circuits in \textit{Hardware
  Description Languages} (HDLs) such as VHDL and (System)
Verilog. With these the designer specifies the behaviour of all the
circuit's logic gates and flip-flops. The result is a
\textit{Register-Transfer Level} (RTL) design, so called because it
models the register-to-register transfer the circuit's signals. A
synthesizer takes the RTL design and produces a low-level
representation (netlist) of it which can be lowered further to create
an ASIC or transformed into an FPGA bitstream. The synthesizer's job
is far from straight-forward and it must -- among other things --
place every component of the circuit on a two-dimensional grid and
ensure that it can operate with the desired clock frequency.

While HDLs offer a great deal of control over the resulting circuit,
they are low-level and lack support for many high-level programming
constructs so using them can be tedious and error-prone. It often
makes sense to work in a higher-level language instead. That workflow
is called High-Level Synthesis (HLS) and is supported by many
tools. For example, Intel's FPGA software generates synthesizable
Verilog code from designs coded in the imperative C-like language
OpenCL.

The main worry of using HLS is that the hardware will not be as
efficient as if an HDL had been used. As the designer works in an
imperative, high-level language their view of the hardware may be
obscured and not easy to visualize. Furthermore, the machine-generated
HDL generated by HLS tools is often opaque and near impossible to
understand. These fears may be unfounded, however, as many studies
have found that HLS does not degrade performance and sometimes even
improve it \citep{lahti2018}.

% Above para not so great...

\subsection{OpenCL}
\label{sec:ocl}
\begin{figure}
\begin{lstlisting}[style=opencl]
__kernel void mul_sd(
    __global float *A,
    __global float *B,
    __global float *C,
    int N) {
    for (uint i = 0; i < N; i++)
        C[i] = A[i] * B[i];
}
__kernel void mul_nd(
    __global float *A,
    __global float *B,
    __global float *C,
    int N) {
    uint i = get_global_id(0);
    if (i < N)
        C[i] = A[i] * B[i];
}\end{lstlisting}
\caption{Single and multiple work item OpenCL kernels for element-wise vector product}
\label{lst:kernels}
\end{figure}

\noindent In 2009 the Kronos Group published the first version of the
OpenCL (Open Computing Language) standard \citep{munshi2009}. The goal
was to replace all vendor-specific languages and application
programming interfaces (APIs) with a common and portable standard for
writing high-performance code across all kinds of
devices. An OpenCL application (unlike one written in a competing
technology such as CUDA) can run \textit{unmodified} on any device
with a compliant OpenCL implementation. Today OpenCL now runs on
millions of CPUs, GPUs, DSPs, and other computing devices.

The OpenCL standard consists of three parts; a specification for a
programming language, a host API, and a device API. The host API runs
on the user's main computing device (e.g., PC) and controls their
accelerator (e.g., GPU), whose functionality is accesible through the
device API. Thus, OpenCL prescribes both how the programmer
should program the accelerator and how to manage it from an external
device.

The OpenCL language closely resembles C. It supports functions, loops,
multiple variable scopes, aggregate data types, and many other
programming constructs familiar to C programmers. A big difference
from C is that OpenCL comes with three explicit pointer spaces;
global, local, and private. These grant the programmer fine-grained
control over data storage. What constitutes global, local, and
private memory is device-specific. Generally the largest and
slowest memory space is global, while the smallest and fastest is
private. Global memory may be sized in gigabytes, while private memory
may only be a few kilobytes. OpenCL has native support for SIMD
types to make it easier to write algorithms for parallel
hardware. Unlike C, OpenCL specifies the bit width of most builtin
types -- an \verb!int!  is always 32 bits and a \verb!long! 64 bits.

OpenCL was designed with massively parallel architectures in mind
(e.g., GPUs) and has builtin support for concurrency in the form of
\textit{work items}. A work item is a small indivisible unit of work
with, ideally, no dependencies to other work items. The OpenCL runtime
can schedule independent work items to maximally exploit the
hardware's potential. Consider the kernels in listing
\ref{lst:kernels} for computing the element-wise vector
product. Launching the first kernel -- \verb!mul_sd! -- causes OpenCL
to instantiate one kernel on one core which runs $N$ iterations of the
for loop. Launching the second kernel, however, causes OpenCL to
instantiate up to $N$ kernels, each of which runs on an available core
and computes one iteration of the for loop. Since the second kernel
runs more computations in parallel, it likely runs much
faster. Sometimes the algorithm's work cannot feasibly be separated
into completely independent work items. For such situations, work
items that must be synchronized can be organized into \textit{work
  groups} which share local memory. However, synchronization is
impossible between work groups.

\subsection{OpenCL on Intel FPGAs}
\label{sec:ocl-fpga}
\begin{figure}
  \begin{lstlisting}[style=opencl]
for (uint i = 0; i < N; i++) {
    B[i] = h(g(f(A[i])));
}\end{lstlisting}
  \caption{A loop-nest that could benefit from pipeline
    parallelism. The functions \texttt{f}, \texttt{g}, and \texttt{h} are
    assumed to be short and inlineable.}
  \label{lst:para}
\end{figure}
\begin{figure}
  \begin{lstlisting}[style=opencl]
channel uint jobs
  __attribute__ ((depth(512)));
__kernel void consumer() {
  while (true) {
    uint job =
      read_channel_intel(jobs);
    if (!msg)
      break;
    process_job(job);
  }
}
__kernel void producer(uint N)
  for (uint i = 0; i < N; i++) {
    uint job = create_job(i);
    send_channel_intel(job);
  }
  send_channel_intel(0);
}\end{lstlisting}
  \caption{Producer-consumer kernels communicating via a channel}
  \label{lst:chans}
\end{figure}

\noindent As FPGAs work very differently from processor-based
hardware, Intel's OpenCL implementation for FPGAs differs in important
ways from other OpenCL implementations. One major difference concerns
parallelism. CPUs and GPUs have multiple cores to run multiple
computations in parallel. Thus, the workload of an OpenCL program
organized into multiple work items can be mapped onto multiple
cores. FPGAs do not have any cores in the usual sense and it is
difficult for Intel's OpenCL FPGA compiler to synthesize code
structured around work items into efficient FPGA bitstreams. Instead,
Intel recommends designers to structure OpenCL code as single-work
item kernels, and to derive most of the parallelism from executing
multiple loop iterations concurrently; a type of concurrency known as
``pipeline parallelism''. For example, assume the three functions
\verb!f!, \verb!g!, and \verb!h! in listing \ref{lst:para} are short,
side-effect free, and represent computations that can be performed in
less than one clock cycle. The FPGA compiler can create a three-stage
pipeline for the loop, where combinational circuits for \verb!f!,
\verb!g!, and \verb!h! constitute the pipelines three stages. This
pipeline can process three iterations of the loop in parallel; while
the \verb!h!-circuit processes data for the first, the
\verb!g!-circuit processes data for the second, and the
\verb!f!-circuit for the third iteration, and so on. While the latency
of every iteration is three cycles, the throughput (initiation
interval) is only one cycle and the latency for the loop as a whole is
$N + 2$ since it takes 2 cycles to fill the pipeline. Though, this
assumes that the latency of every function is fixed and
predictable. If one function executes operations with variable
latencies, such as memory accesses, the pipeline may need to
stall. Moreover, pipeline parallelism cannot increase the throughput
of the loop beyond one cycle -- for that one has to use other
techniques.

The FPGA compiler supports many compiler directives
(``pragmas'') that can help it better optimize difficult
loops or loops containing invariants it can not
prove on its own. Two important directives are \verb!#pragma ivdep!
and \verb!#pragma unroll N!. The first tells the compiler that the
loop contains no loop-carried dependencies and therefore it
can reorder the loop iterations as it pleases. The second tells it to
duplicate the loop body $N$ times. This means that the resulting
circuit will use $N$ times as many gates, but, potentially, run $N$
times faster. Other compilers ignore these directives.

The FPGA compiler extends OpenCL with syntax for declaring channels
for inter-kernel communication. Channels resemble the pipes feature in
OpenCL 2.0 which the FPGA compiler does not support. Channels are
implemented in on-chip memory as first in first out (FIFO) buffers of
the desired depths and are very fast. \Cref{lst:chans} shows a simple
producer-consumer example, where one kernel calls \verb!send_channel_intel!
to put jobs on the FIFO and the other calls \verb!read_channel_intel!
to remove them. Both functions block if the FIFO is full or empty.

\section{SNN Simulation}
\label{sec:ssn-sim}

\begin{figure}
  \begin{lstlisting}[style=pseudocode]
for t in range(n_tics):
    for i in range(n_neurons):
        N[i] = update_neuron(N[i])
        if spikes(N[i]):
            Q = enqueue(Q, i, t)
    for i in range(n_neurons):
        c = collect(Q, i, t)
        N[i] = update_psc(Q, c)
        Q = dequeue(Q, i, t)\end{lstlisting}
  \caption{Synchronous SNN simulation}
  \label{lst:basic}
\end{figure}

\noindent Having reviewed SNNs, the Potjans-Diesmann microcircuit and
our implementation tools, we now discuss methods for simulating SNNs
efficiently. We explore SNN simulation in general before introducing
the ideas and algorithms we use to optimize our simulators. The end
result is a taxonomy of twelve simulator families, grouped by their
algorithms and implementation styles. We illustrate most of our points
with Pythonesque pseudo-code, extended with two keywords;
\verb!parfor! and \verb!atomic!. The former for loops whose iterations
are independent and therefore can be executed in parallel. The latter
for operations that must execute as one indivisible unit. Text in
\verb!{brackets}!  explains what the pseudo-code should do at that
point.

Broadly speaking, SNN simulation can be categorized on whether it is
synchronous (time-stepping) or asynchronous
(event-driven). Synchronous simulation updates the state of every
simulated object at every tick of a clock, regardless of whether it is
necessary or not \citep{brette2007}. Asynchronous simulation only
updates simulated objects whey they receive external stimuli, i.e.,
events. In an asynchronous SNN simulation, spike emission and
reception constitute the events, since the state of a neuron at a time
between two events can be calculated easily (and generally is
unimportant). Hybrid strategies, with asynchronous updates for some
parts of the SNN and synchronous updates for other parts, are
possible.

For SNN simulation, asynchronicity offers precision advantages. The
neuron's membrane potential only has to be recomputed when it receives
spikes. If this happens only rarely the simulator can afford to use
more sophisticated methods than (repeated) forward Euler to solve the
LIF equation (\cref{eqn:lif}). Also, spikes can be sent and received
at any time and do not have to be confined to a discrete grid. For
example, a synchronous simulator with a 0.1 ms time steps may not be
able to represent spikes sent at times that are not multiples of 0.1
ms. Asynchronicity may also have scalability advantages as the
neurons' states do not have to be synchronized to a global
clock. However, asynchronicity entails irregular computation and
irregular memory accesses -- traits that are extremely undesirable on
modern hardware. Furthermore, dense SNNs have many more synapses than
neurons which results in cascading effects. One spiking neuron ``wakes
up'' thousands of its neighbours, causing them to spike and in turn
wake up thousands of their neighbours. On the whole, the event
processing overhead may dominate over whatever computational savings
not being bound by a global clock brings. For example,
\cite{pimpini2022} presents a sophisticated asynchronous CPU-based SNN
simulator that supports speculative execution so that future neuron
states can be computed in advance and then rolled back if received
spikes (``stragglers'') invalidates their predicted state. While the
authors measured accuracy improvements, the performance was
lackluster. For these reasons, most high-performance SNN simulators,
including ours, are synchronous.

Synchronous simulation splits the simulation task into two phases; one
for updating neurons and one for transferring spikes. Listing
\ref{lst:basic} shows the basic algorithm. It uses two data
structures; an indexable data structure, \verb!N!, to stores the state
of all neurons and a queue, \verb!Q!, to keeps track of spikes in
flight. For every neuron the algorithm calls \verb!update_neuron! to
update its membrane potential, presynaptic current, and refractoriness
in accordance with \cref{eqn:mem,eqn:psc,eqn:ref}. If \verb!spikes!
indicates that the neuron spikes, it enqueues the neuron's index $i$
and the current time step $t$ in \verb!Q!. In the next phase (line 6
to 9) the algorithm calls \verb!collect! on the queue to aggregate
current destined to the $i$:th neuron at time step $t$. The call to
\verb!update_psc! adds the aggregated current to the neuron's
presynaptic current. Finally, the algorithm removes the current it
just handled from the queue. In this listing we include the
\verb!for t in range(n_tics)! loop which shows that the algorithm
repeats the two phases \verb!n_tics!  times. However, for brevity's
sake, we omit this outer loop in the following listings.

The work required for updating the neurons' state for one tick is in
the order of $O(N)$, where $N$ is the number of neurons --
i.e. linear. However, for transferring spikes it is $O(f\Delta
tpN^2)$, where $p$ is the probability that two randomly chosen neurons
are connected by one or more synapses, $f$ the average spiking rate,
and $\Delta t$ the tick duration. So the transfer time is proportional
to the network's density and quadratic in $N$ -- i.e. for large $N$ it
dominates. Furthermore, updating the membranes and presynapses
require a handful of multiplications per neuron -- cheap on modern
hardware -- whereas spike transfer requires expensive reads and writes
to and from non-contiguous memory. Hence, we will focus on the
transfer phase which is where synchronous simulators spend most of
their time in the remainder of this section.

\subsection{Pushing and Pulling}
\label{sec:push-pull}
\begin{figure}
  \begin{lstlisting}[style=pseudocode]
parfor i in range(n_neurons):
    spikes = False
    if R[i] == 0:
        x = p22*U[i] + p21*I[i]
        spikes = x >= U_thresh
        if spikes:
            x = 0
            R[i] = t_ref_tics
        U[i] = x
    else:
        U[i] = 0
        R[i] -= 1
    if spikes:
        parfor j, d, w in syns_from(i):
            atomic W[t + d, j] += w
    I[i] = p11*I[i] + T[t, i]*wpsn
    I[i] += W[t, i]\end{lstlisting}
  \caption{One time step of push-based spike transfer}
  \label{lst:push}
\end{figure}

\begin{figure}
  \begin{lstlisting}[style=pseudocode]
parfor i in range(n_neurons):
    A[t, i] = False
    if R[i] == 0:
        x = p22*U[i] + p21*I[i]
        if x >= U_thresh:
            A[t, i] = True
            x = 0
            R[i] = t_ref_tics
    else:
        x = 0
        R[i] -= 1
    U[i] = x
    s = 0
    for j, d, w in syns_to(i):
        if A[t - d, j]:
            s += w
    I[i] = p11*I[i] + T[t, i]*wpsn + s\end{lstlisting}
    \caption{One time step of pull-based spike transmission}
    \label{lst:pull}
\end{figure}

\noindent Researchers have identified ``pushing'' and ``pulling'' as
two general strategies for designing algorithms for graph problems
(\cite{besta2017}; \cite{grossman2018}; \cite{ahangari2023}). A
push strategy transfers signals \textit{from a node to its neighbours}
by writing to the neighbours incoming signal buffers. In contrast, a
pull strategy transfers signals \textit{to a node from its neighbours}
by sweeping through the node's neighbours and checking whether they
have any signals to be delivered to the node. I.e., the receiver node
has to ``go and ask'' its neighbours whether they have signals for
them. SNN simulation is a graph problem involving node-to-node
transfer of signals and it too can be characterized in terms of
pushing and pulling. Listing \ref{lst:push} and \ref{lst:pull}
illustrate the two strategies. In both listings, the values of the
scalars \verb!p11!, \verb!p21!, and \verb!p22! come from
\cref{eqn:p11p22,eqn:p21}. The arrays \verb!U!, \verb!I!, and \verb!R!
contain the neurons' membrane potentials, presynaptic potentials, and
refractory counters. When the neuron's membrane potential exceeds
\verb!U_tresh! the neuron spikes and becomes refractory for
\verb!t_ref_tics! tics. The expression \verb!T[t, i]!  denote the
number of thalamic spikes received by the $i$:th neuron at the $t$:th
time step.

The strategies differ in how they transfer spikes. The push strategy
transfers them when \verb!spikes! indicates that a neuron spikes. It
calls \verb!syns_from! to fetch an iterator of all synapses
\textit{originating} from the $i$:th neuron. The three-tuples $(j, d,
w)$ it retrieves represent the synapses; $j$ the index of the
destination neuron, $d$ the delay in time steps, and $w$ the
current. For every three-tuple it writes to an element of \verb!W!, a
two-dimensional array buffering current to be delivered. The element
\verb!W[t, i]!  is the amount of presynaptic current the $i$:th neuron
receives at the $t$:th time step. The \verb!syns_to! call in listing
\ref{lst:pull} works exactly like the \verb!syns_from! call, except it
returns all synapses \textit{terminating} at the $i$:th neuron and $j$
in the three-tuples $(j, d, w)$ denotes the \textit{originating}
neuron. Array elements \verb!A[t, i]! indicate whether the $i$:th
neuron spikes at time $t$ or not. The size of the longest synapse
delay is in practice bounded and small so both \verb!A! and \verb!W!
can be implemented as statically sized wrap-around arrays -- a
technique we cover in \cref{sec:sizing}. Note that every neuron can be
handled independently so we use the \verb!parfor! construct for both
algorithms' outer loops.

Pushing of synaptic current to their neighbours happens on line 16 and
17 of the push strategy's listing. Neurons spike infreqently, but when
they do, the algorithm must retrieve all their synapses and write
their current to the array \verb!W!. This part of the algorithm is
performance-critical. First, the algorithm accesses all the neuron's
synapses which can be expensive, even if they are stored in contiguous
memory. Second, the algorithm gathers and scatters data from and to
uncorrelated indices of \verb!W!. These are costly operations since
the accesses to \verb!W! cannot be coalesced or cached.\footnote{The
memory addresses are too far apart for any coalescing or caching
scheme to be efficient.} To make matters worse, at one time step
multiple neurons can write to the same indices of \verb!W!. I.e.,
\verb!W[t + d, j] += w!  has to be executed atomically to avoid data
races. As the model is densely connected, races are not uncommon and
should be accounted for. Partition-awareness, as suggested by
\cite{besta2017}, is not an option because of the density and
randomness of the connections, making most of them remote and not
local.

\begin{figure}
  \begin{lstlisting}[style=pseudocode]
for i in range(n_neurons):
    {Update neuron state as before}
    if spiked:
        Q = enqueue(Q, i)
for i in contents(Q):
    parfor j, d, w in syns_from(i):
        W[t + d, j] += w
Q = clear(Q)\end{lstlisting}
\caption{Deferred push-based spike transfer}
\label{lst:push2}
\end{figure}

The pull-based algorithm instead has the receiver neurons responsible
or ``pulling in'' current. Spiking merely sets an element in \verb!A!
to true and does not trigger current transfer. On subsequent time
steps, neurons connected to that neuron checks if it spiked at time
$t-d$ and, if so, adds its synaptic current. The upside of this
algorithm is that it is synchronization-free; neurons do not write to
shared memory. The major drawback is that every neuron at every time
step must check all its incoming synapses to see from which of them it
receives current. Additionally, the algorithm reads from scattered
memory on line 18. The large amount of data it reads probably makes
it inefficient, unless the number of computational units is
large and memory reads are substantially cheaper than writes. Neither
of which is true for our FPGA. Consequently, we focus on push-based
spike transfer in this work.

Listing \ref{lst:push2} shows a variant of push-based spike transfer
that first collects spiking neurons and then transfers their synapses
current in a dedicated phase. Collection can either be done with a
marking array, at the expensive of wasting memory, or with a queue (as
in the listing), at the expense of making loop iterations
dependent. The choice depends on the target platform. Though,
splitting the update and spike transfer into two phases reduces the
number of cache conflicts which is advantagenous. Lines
13 to 15 of the basic push algorithm can flush prefetched and cached
parts of the \verb!U!, \verb!I!, and \verb!R!  arrays.

\subsection{Buffer Sizing and Wrapping}
\label{sec:sizing}

To reduce the size of the \verb!W! array we use ``wrap-around
indexing''. The indices of the rows in \verb!W! written to in one time
step lies within the interval $t + 1$ to $t + d_\mathrm{max} - 1$,
where $d_\mathrm{max} - 1$ is the network's largest synaptic delay and
is small. Moreover, at time step $t$ rows $0$ to $t - 1$ will not be
read again so that space can be reused. We do that by setting the
number of rows in \verb!W! to $d_\mathrm{max}$ and use modular
arithmetic to index rows. The expressions for accessing \verb!W! on
line 15 and 17 in listing \ref{lst:push} become
\verb!W[(t + d) % D_MAX, j]! and \verb!W[t % D_MAX, i]!,
respectively. We choose a large enough value for $d_\mathrm{max}$ by
evaluating the cumulative distribution function of the Gaussians we
sample the slower excitatory synapses delays from, $\mathcal{N}(1.5,
0.75)$:
\begin{equation}
  P(\mathcal{N}(1.5, 0.75) \leq 6.4) \approx 0.999999999968:
\end{equation}
This shows that with $\Delta t = 0.1$, 64 rows is more than enough
since the probability of sampling a synaptic delay longer than 6.4 ms
is astronomically low. It is also a power of two so we use masking to
realize the modular arithmetic. With this scheme we also have to clear
\verb!W[t, i]!  after reading it to avoid double reads.

Even with only 64 rows and assuming four-byte floats, the \verb!W!
array still consumes $64\cdot4\cdot 77169 \approx 20$ megabytes which
is more than we can fit in on-chip memory. We could store the array in
off-chip memory -- which is plentiful -- or use half-precision two-byte
floats instead. Neither solution is satisfactory. As we argued in
\cref{sec:push-pull}, we need fast reads and writes to uncorrelated
addresses of \verb!W!  which off-chip memory doesn't give us. We also
prefer not to lower the numeric the precision as that makes the
simulation less accurate. A third option is to store all spiking
neurons in a queue and only activate a subset of their synapses at a
given time step.

\subsection{Just-In-Time Spike Transfer}
\begin{figure}
  \begin{lstlisting}[style=pseudocode]
parfor i in range(n_neurons):
    {Update U[i], R[i] as before}
    spiked[i] = {True if neuron spiked}
    I[i] = p11*I[i] + T[t, i]*wpsn
for rt in range(D_MAX):
    delay = (t - rt) % D_MAX
    if delay < D_MAX - 1:
        for n in enqueued_at(Q, rt):
            syns = syns_from(n, delay)
            parfor j, d, w in syns:
                I[j] += w
    else:
        Q = dequeue(Q, rt)
rt = t % D_MAX
for i in range(n_neurons):
    if spiked[i]:
        Q = enqueue(Q, rt, i)\end{lstlisting}
  \caption{Three-phase just-in-time spike transfer}
\label{lst:jit}
\end{figure}

\noindent By activating all the neuron's synapses at once, the push
algorithm works harder than necessary. Obviously, all synapses must
\textit{eventually} be activated, but \textit{right now} only synapses
with a delay of one time step must be activated since the current they
transfer will be read at the next time step. This observation leads us
to a ``lazy'' just-in-time algorithm which keeps all spiking neurons
in a queue for a fixed number of time steps. At every time step it
activates those synapses whose current is read at the next time
step. Thus, a neuron that spikes at time $t$ will have its
one-tick-delay synapses activated at time $t + 1$, its two-tick-delay
synapses at time $t + 2$, and so on. Listing \ref{lst:jit} sketches a
three-phase algorithm built on this idea. The first phase (line 1 to
4) updates \verb!U! and \verb!R! as before and marks spiking neurons
in \verb!spiked!. The third phase (lines 14 to 17) enqueues the marked
neurons with a ``relative timestamp'', \verb!rt!. The second phase
(lines 5 to 13) scans the queue and calls \verb!enqueued_at! to fetch
previously enqueued neurons. The \verb!syns_from! function works as
before, but now has a second parameter to select synapses with the
given delay. Suppose that the simulator simulates time step 100 and
that \verb!rt! is 10 in one iteration of the loop. Since $26 \equiv
(100 - 10) \mod d_{\mathrm{max}}$ ($d_{\mathrm{max}}=64$) all synapses
with 26 time steps of delay of queued neurons with a relative
timestamp of 10 will be activated. These neurons ought to have been
stored at time step $t = 100 - 26 = 76$ which is indeed the case since
$10 \equiv 74 \mod d_\mathrm{max}$. After the neurons have stayed in
the queue for \verb!D_MAX - 1! time steps, the \verb!dequeue! call
evicts them. Though with a sensible FIFO implementation eviction is a
no-op so we omit it in future pseudo-code.

How much memory do the just-in-time algorithm save? Through
experimentation we found that the average number of spiking neurons
per time step is 23, meaning that the expected number of
elements in \verb!Q! is $23\cdot d_{\mathrm{max}}=1472$. We generously
round it up to 4096 and, as neuron indices take four-bytes to store,
the total size of the queue is 16 kilobytes, which comfortably fits in
on-chip memory.\footnote{As in \cref{sec:sizing}, we could use the
cumulative distribution function to show that 4096 elements is enough
to make the risk of overflow virtually zero.}

To improve the just-in-time algorithm we use ``lanes''. Essentially,
we duplicate the \verb!W! array so that synapses of multiple spiking
neurons can be activated simultaneously. The algorithm writes the
synaptic current of the first neuron it handles in the first lane, of
the second neuron in the second lane, and so on. Synaptic current of
neurons in different lanes does not interfere. The update phase has to
be adjusted accordingly and must sum the incoming current from all
lanes. As each lane consumes about 320 kb of memory we can fit at most
16 lanes.

\subsection{Horizon-Based Spike Transfer}
\begin{figure}
  \begin{lstlisting}[style=pseudocode]
parfor i in range(n_neurons):
    I[i] += W[t % H, i]
    W[t % H, i] = 0
    {update U[i] and R[i] as before}
    I[i] = p11*I[i] + wspn*T[t, i]
for i in range(D_MAX / H):
    rt = (t - H*i - 1) % D_MAX
    d_from = H*i + 1
    d_to = d_from + H
    for n in enqueued_at(Q, rt):
        syns = syns_from(
            n, d_from, d_to)
        parfor j, d, w in syns:
            W[(d + t) % H, j] += w
rt = t % D_MAX
for i in range(n_neurons):
    if spiked[i]:
        Q = enqueue(Q, rt, i)\end{lstlisting}
  \caption{Three-phase spike transfer with configurable horizon}
  \label{lst:horizon2}
\end{figure}

\noindent The just-in-time algorithm requires a fair amount of
bookkeeping. It activates the spiking neuron's synapses
$d_{\mathrm{max}}$ times instead of just once, and the loop on line 10
and 11 iterates many fewer times than the corresponding loop on line
14 and 15 of the basic push algorithm. This loop is an important
source of parallelism and running it many times with fewer iterations
is much worse for performance than running it fewer times with many
iterations. Our solution is to reintroduce a smaller version of the
spike buffer whose number of rows, $h$ is a factor of $d_\mathrm{max}$
so that when a neuron spikes we write to the buffer $d_\mathrm{max} /
h$ times. Listing \ref{lst:horizon2} shows the concept. The
\verb!syns_from! function now retrieves synapses of the neuron whose
delay is within the range $d_\mathrm{from}$ to $d_\mathrm{to} -
1$. Suppose $d_\mathrm{max}=64$, $t=100$ and $h=16$. The loop on lines
5 to 13 iterates four times, the relative timestamps assumes the
values 35, 19, 3, and 51, and the half-open intervals the values [1,
  17), [17, 33), [33, 49), and [49, 65). Thus, the inner loop on line
        10 to 14 activates all synapses with the given delays of the
        neurons stored at the given relative timestamps. With this
        scheme we trade-off on-chip memory for better concurrency.

Note that we add the current from the spike buffer to the presynaptic
potential on line 2, before we update the membrane's potential and
we add one to \verb!d_from! on line 7. This is necessary since the
algorithm transfers spikes one time step later than the basic
push algorithm.

\subsection{Storing Synapses}
\label{sec:syn-store}
\begin{figure}
  \begin{lstlisting}[style=pseudocode]
def syns_from(n, d_from, d_to):
    start = X[n, d_from]
    end = X[n, d_to]
    for i in range(start, end):
        yield S[i]\end{lstlisting}
\caption{Indexed access to synapse data}
\label{lst:index}
\end{figure}

\noindent Previous sections' pseudo-codes imply that it is very
important that synapses can be queried by their sender neuron
quickly. In particular, synapses should be stored so that one
\verb!syns_from! call only accesses memory in one contiguous chunk. We
fullfill these goals by storing the synapses as an array sorted on
sender neuron, receiver neuron, and delay that we query with a
prebuilt index keyed on sender neuron and delay. This means that
finding all synapses for a particular neuron or neuron-delay
combination requires only two index lookups; one for the first synapse
and one for the last synapse. Moreover, as the index implicitly stores
the sender neuron, we only need eight bytes to represent a synapse;
four for the single-precision weight (32 bits), three for the
destination neuron's id (17 bits), and one for the synaptic delay (6
bits).\footnote{With just-in-time transfer the delay is also stored
implicitly.} Listing \ref{lst:index} shows how \verb!syns_from!
performs index look-ups. The two-dimensional array \verb!X!
represents the index and \verb!S! the synapse array so that
\verb!X[n, d_from]! contains the index in \verb!S! where the first
synapse of neuron \verb!n! with delay \verb!d_from! is
stored. Ideally, the synapses should also be prefetched.

\begin{figure}
  \begin{lstlisting}[style=pseudocode]
parfor c in range(N_CLS):
    for i in contents(Q):
        syns = syns_from(i, c)
        for j, d, w in syns:
            W[t + d, j] += w
Q = clear(Q)\end{lstlisting}
\caption{Spike transfer with partitioned synapses}
\label{lst:parts}
\end{figure}

\subsection{Disjoint Synapses}
\label{sec:disjoint}

A headache for push-based spike transfer is the data race caused by
multiple synapses delivering current received by the same neuron at
the same time step. This is why we can't use \verb!parfor! on line 5
of listing \ref{lst:push2}, for example. We cannot completely solve
this problem, but we can alleviate it by partitioning the synapses
into disjoint \textit{classes}, so that synapses of different classes
never trigger writes to the same memory addresses at the same
time. Then every class can be handled by a separate thread. The
pseudocode in listing \ref{lst:parts} modifies the transfer phase of
the deferred push algorithm from \ref{lst:push2} to exploit of this
idea.\footnote{We can of course us the same technique to improve all
other algorithms we have discussed.}  The constant \verb!N_CLS!
denotes the number of synapse classes and the extra argument to
\verb!syns_from! which synapse class to query. Many ways of
partitioning the synapses into disjoint classes are possible. For
example, by delay so that one thread writes one-time step synapses,
the next thread two-time steps synapses, and so on. Another by
destination neuron so that one thread handles synapses going to
neurons whose index is between 0 and 199, another those
between 200 and 399, and so on.

\begin{figure}
  \begin{lstlisting}[style=pseudocode]
o0 = X[i][0]
o1 = X[i][D_MAX]
parfor c in range(N_CLS):
    for o in range(o0 + c, o1, N_CLS):
        j, d, w = S[o]
        W[(t + d) % D_MAX, j] += w\end{lstlisting}
  \caption{Spike transfer over interleaved synaptic storage.}
  \label{lst:parts2}
\end{figure}

We choose to partition by the destination neuron's congruence
class. The method has low overhead and evenly distributes the synapses
over the classes since the least significant bits of the neuron index
is almost random. To retain the contiguous storage we interleave the
synapses. That is, if a neuron's synapses are found at indexes $o_0$
to $o_1 - 1$, then all synapses of class $c$ are stored at indices
$o_0 + n_ci + c$, where $n_c$ is the number of classes and $i$ is a
non-negative integer. Interleaving synergizes with banked memory
common on many GPUs. On our FPGA, it means that we can have
conflict-free dedicated memory ports for every synapse class. The
method causes some memory waste however. For example, if there are
four classes and all destination neuron indices of all synapses of
some neuron happen to be congruent with $2 \mod 4$, then all indices
other than $o_0 + 4i + 2$ will be vacant. I.e., 75\% of the space will
go to waste. In general, the memory consumption for storing a neuron's
synapses grows from $n_ss$, where $n_s$ is its number of synapses and
$s$ the synapse size to $n_cls$, where $l$ is the number of synapses
in the neuron's largest class. In practice, the memory waste
is manageable; in the order of 5-30\% depending on the number of
classes and horizon. The more the classes and the shorter the horizon
the more uneven the classes become and the more waste. We do not mark
vacancies and instead fill them with synapses carrying no current and
terminating at idempotent neurons. This way, the code in listing
\ref{lst:parts2} does not need to check whether the index is vacant.

\subsection{More on Parallelism}
\label{sec:more-para}
\begin{figure*}
  \begin{subfigure}{0.48\textwidth}
    \begin{lstlisting}[style=pseudocode]
parfor i in range(n_neurons):
    {Update neuron state as before}
    if spikes:
        write(to_transfer, i)
write(to_update, DONE)
read(to_update)\end{lstlisting}
    \caption{Update kernel}
  \end{subfigure}
  \begin{subfigure}{0.48\textwidth}
    \begin{lstlisting}[style=pseudocode]
while True:
    i = read(to_transfer)
    if i == DONE:
        break
    parfor j, d, w in syns_from(i):
        W[t + d, j] += w
        write(to_update, True)\end{lstlisting}
    \caption{Transfer kernel}
  \end{subfigure}
  \caption{Basic spike transfer using two kernels}
  \label{lst:multi}
\end{figure*}

\noindent Our algorithms' source of parallelism is the \verb!parfor!
keyword. Iterations of such loops are independent and can be executed
concurrently in duplicated hardware. The replication is realized
differently on different targets. On CPUs we use SIMD and on GPUs the
single-instruction multiple threads (SIMT) execution model
``automatically'' parallelizes the computation. With Intel's OpenCL
SDK we use the \verb!#pragma unroll N! and \verb!#pragma ivdep!
compiler directives to instruct the compiler to replicate loop
hardware. Essentially, this ``widens'' data paths, allowing more data
to be processed in parallel. But FPGAs can also run more data paths in
parallel, akin to how multiple CPU cores can run multiple threads. We
implement this parallelism by dividing the algorithms over multiple
concurrent communicating kernels.

Listing \ref{lst:multi} restructures the deferred push-based spike
transfer algorithm in \ref{lst:push2} as two kernels. It uses two
blocking FIFOs, \verb!to_transfer! and \verb!to_update!, that the
kernels can \verb!read! and \verb!write! to. When a neuron spikes, the
update kernel sends its index to the spiking kernel which activates
that neuron's synapses. Hence, the neuron update and spike transfer
phases run in parallel. When the update kernel has updated all neurons
it sends a \verb!DONE!  message and waits for a reply from the spike
transfer kernel. When it receives one, it knows that the spike
transfer kernel has transferred all spikes and it proceeds to the next
time step. As inter-kernel communication is not well-supported in
OpenCL, we implement it using the low-latency Intel-specific channel
extension.

\begin{figure*}
    \begin{subfigure}{0.48\textwidth}
      \begin{lstlisting}[style=pseudocode]
for i in range(n_neurons):
    I[i] += read(to_update)
A = []
for i in range(n_neurons):
    {Update neuron state as before}
    if spiked[i]:
        A.append(i)
write(to_transfer, A)\end{lstlisting}
      \caption{Update kernel}
  \end{subfigure}
  \begin{subfigure}{0.48\textwidth}
    \begin{lstlisting}[style=pseudocode]
for i in range(n_neurons):
    write(to_update, W[t % H, i])
    W[t % H, i] = 0
for i in range(D_MAX / H):
    rt = (t - H*i - 1) % D_MAX
    for n in enqueued_at(Q, rt):
        syns = syns_from(n, rt + H)
        parfor j, d, w in syns:
            W[(d + t) % H, j] += w
for n in read(to_transfer):
    Q = enqueue(Q, t % D_MAX, n)\end{lstlisting}
    \caption{Transfer kernel}
    \end{subfigure}
    \caption{Horizon-based spike transfer using two kernels}
    \label{lst:multi-horiz}
\end{figure*}

Listing \ref{lst:multi-horiz} similarily restructures the
horizon-based algorithm as two kernels. Since the \verb!W! array is
stored in local memory, the multi-kernel approach requires an explicit
synchronization step (lines 1 and 2, and 1 to 3) for sending
presynaptic current from the transfer kernel to the update kernel over
the \verb!to_update! channel. After synchronization, the transfer
kernel can freely write synaptic current from queued neurons to
\verb!W!. While updating neurons, the update kernel keeps track of
which of them spiked and sends them to the transfer kernel. The
transfer kernel adds them to its queue \verb!Q!.

\subsection{Implementations}
\label{sec:impl}
\begin{figure}
  \center
  \includegraphics[width=\linewidth]{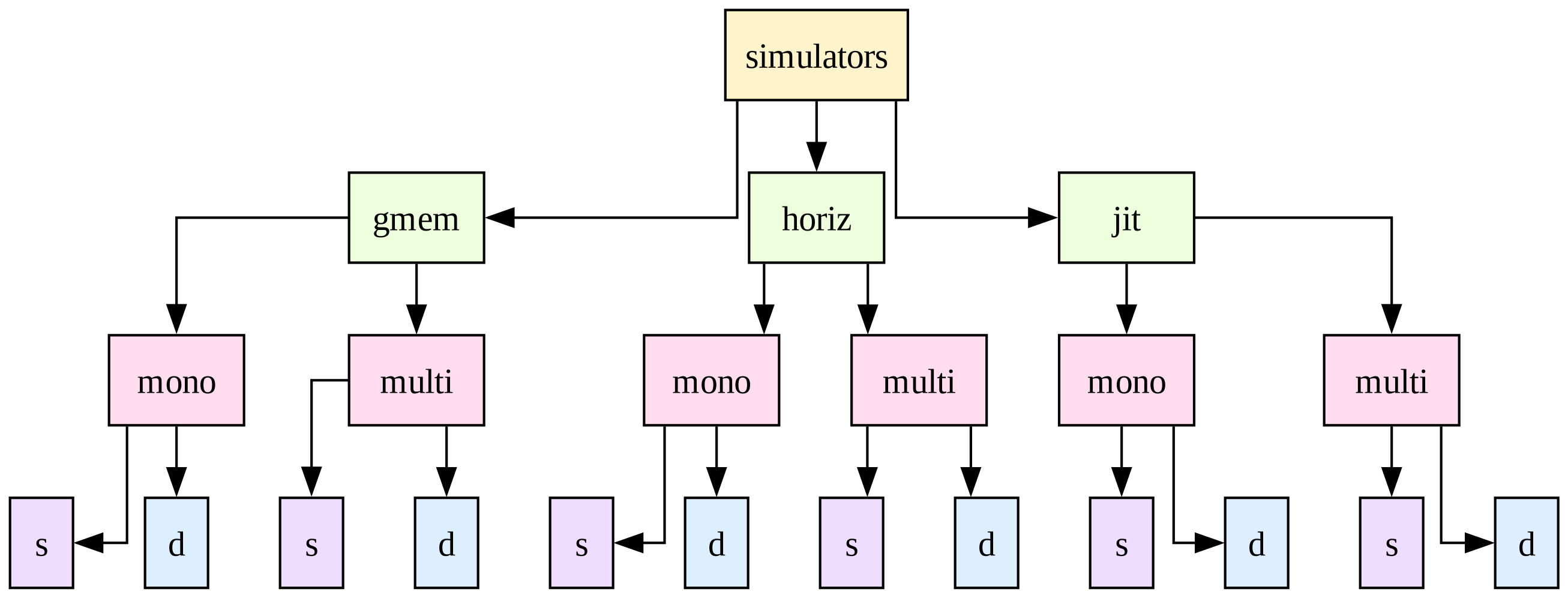}
  \caption{Taxonomy of our simulator implementations}
  \label{fig:taxonomy}
\end{figure}
\begin{figure*}
  \center
  \includegraphics[width=\linewidth]{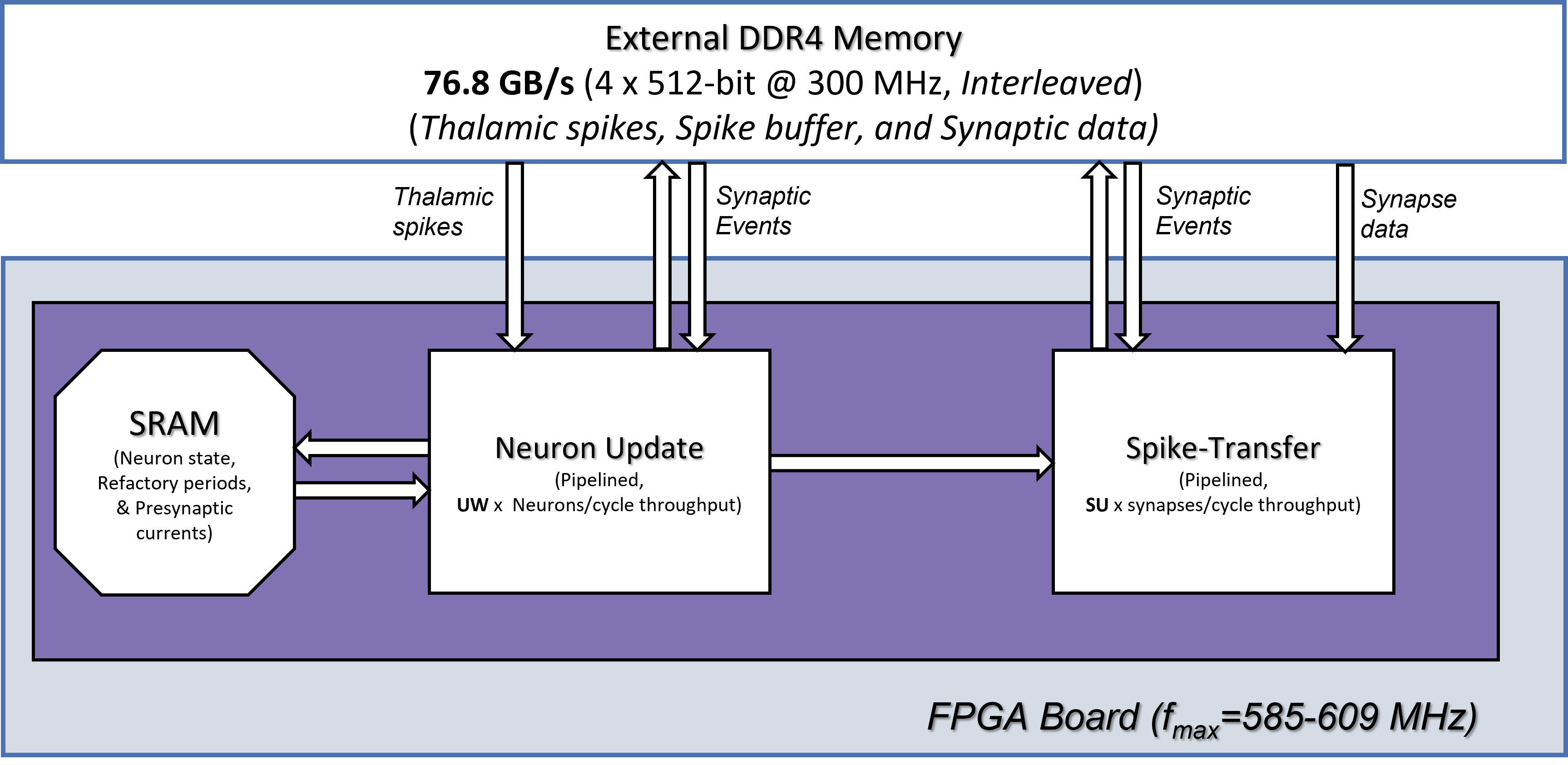}
  \caption{Schematic of the monolithic gmem implementations}
  \label{fig:gmem}
\end{figure*}
\begin{figure*}
  \center
  \includegraphics[width=\linewidth]{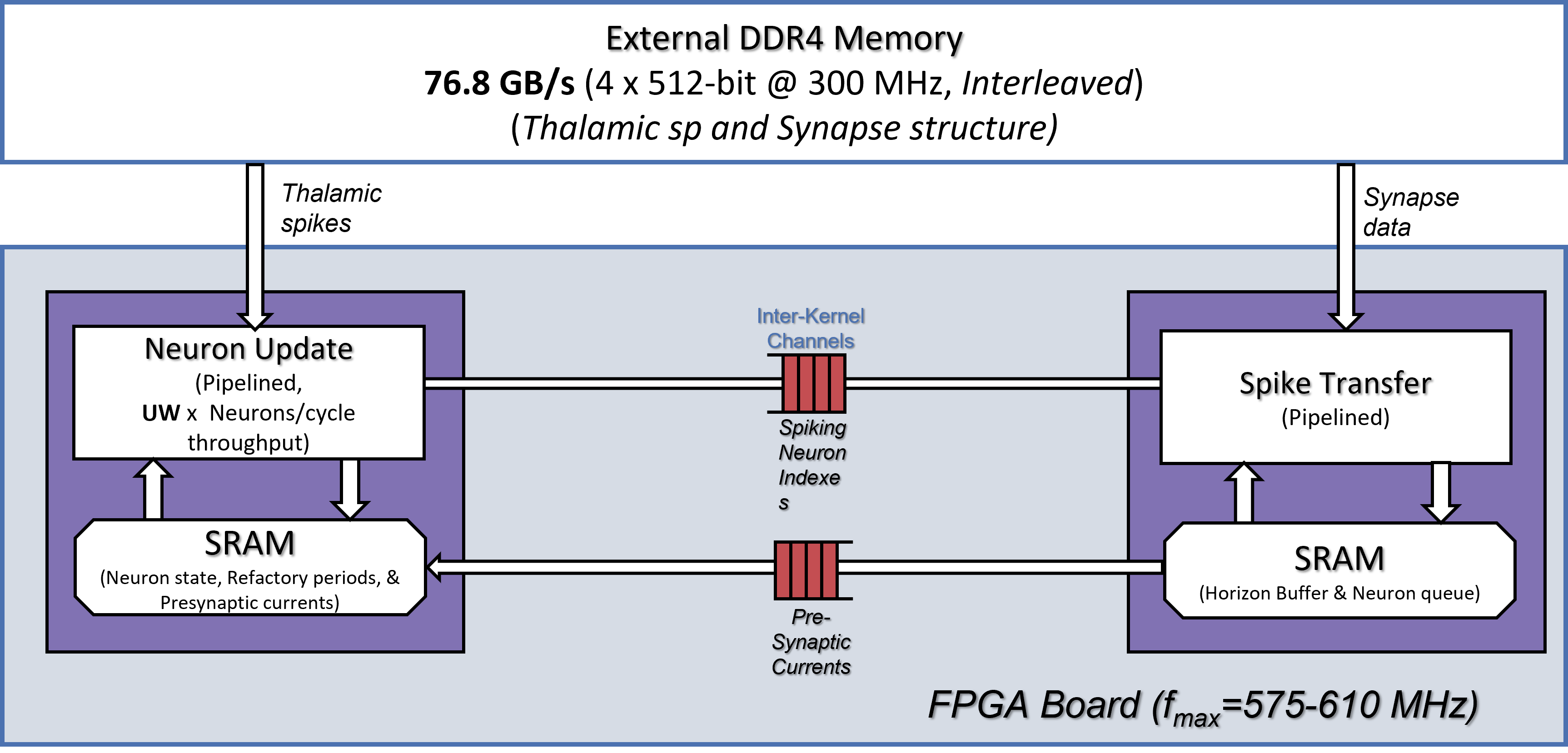}
  \caption{Schematic of the multi-kernel horiz implementations}
  \label{fig:gmem}
\end{figure*}

Given the algorithms and data structures presented, we have created
twelve simulator families arranged into a taxonomy in
\cref{fig:taxonomy}. The taxonomy's second level groups the simulators
on their spike transfer algorithm. It can either be the basic
push-algorithm from listing \ref{lst:push} (gmem), the just-in-time
algorithm from listing \ref{lst:jit} (jit), or the horizon-based
algorithm from listing \ref{lst:horizon2} (horiz). These three
algorithms differ in how eagerly they activate synapses. The gmem
algorithm is maximally eager and activates all the spiking neuron's
synapses at once. Thus, it has to use global off-chip memory to transfer
spikes, hence its name. The jit algorithm is maximally lazy and only
activates synapses whose current should arrive at the next time
step. The horizon algorithm is a mix of the two and activates as many
synapses as its horizon allows. The taxonomy's third level shows
whether the simulator uses multiple communicating kernels (multi) or a
single, monolithic kernel (mono). The last level specifies whether the
simulator uses double (d) or single precision (s) floating point
values for neuron state. This applies to the membrane potentials,
presynaptic currents, and their related coefficients, but not to the
spikes and spike transfer buffers which always are in
single-precision. In the following, we use a notation based on their
taxonomic grouping to refer to the simulators. For example,
``horiz/multi/s'' refers to the horizon-based multi-kernel
single-precision simulators.

All simulators also have parameters for tuning their performance
characteristics. Some are unique to certain families while others are
shared. The \textit{update width} is the most
fundamental one and controls how many neurons the simulator updates at
once. We implement it by unrolling loops such as the loop on line 1 to
4 of listing \ref{lst:jit}. More loop-unrolling is not always better
as it causes the compiler to duplicate the hardware used to synthesize
those loops. The gmem/mono and gmem/multi simulators have a
\textit{synapse unroll} (SU) parameter which controls how many times
their spike transfer loops are unrolled (line 6 and 7 of listing
\ref{lst:push2} and lines 5 and 6 of listing \ref{lst:multi}). More
unrolling allows more synapses to be fetched from memory
simultaneously. The \textit{synapse classes} (SC) parameter determines
the number of congruence classes (see \cref{sec:disjoint}). More
classes allows more synapses to be activated in parallel, but also
increases memory usage. The horiz simulators have a
\textit{horizon length} (H) parameter which determines how many time
steps worth of synapses the simulator activates at once. The jit
simulators have a \textit{lane count} (LC) parameter that controls how
many lanes they uses to transfer current. For convenience and
efficiency, all parameters are positive powers of two.

\section{Results}
\label{sec:results}

We evaluate our simulators on three axes -- correctness, speed, and
energy consumption -- using the same experimental regimen for each. We
synthesize every simulator for our FPGA with Intel FPGA SDK for OpenCL
version 21.2 and use it to simulate ten seconds of biological time
with a time step of 0.1 ms (i.e. 100 000 ticks) on ten different
random microcircuit instances. From these 100 runs we compute mean
values. We run all simulators on the full version (scale 1.0) of the
microcircuit. Due to time constraints, we only synthesize a handful of
simulators for every family and parameter combination. Some simulators
could achieve better performance with a more exhaustive search.

\subsection{Correctness}
\begin{figure*}
  \center
  \includegraphics[width=\linewidth]{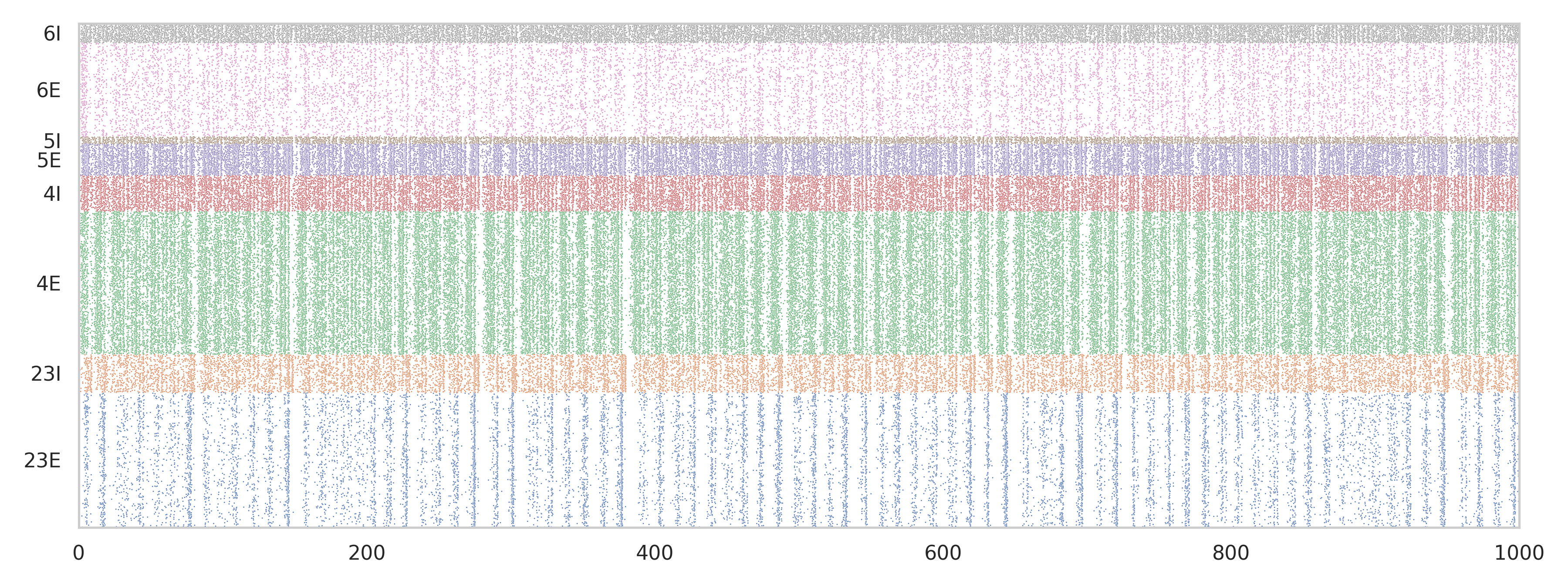}
  \caption{Spike plot of the first 1000 ms of simulation. The
    rhythmic nature of the microcircuit's spiking pattern is apparent.}
  \label{fig:scatter}
\end{figure*}
\begin{figure*}
  \begin{subfigure}{.33\textwidth}
    \centering
    \includegraphics[width=0.9\linewidth]{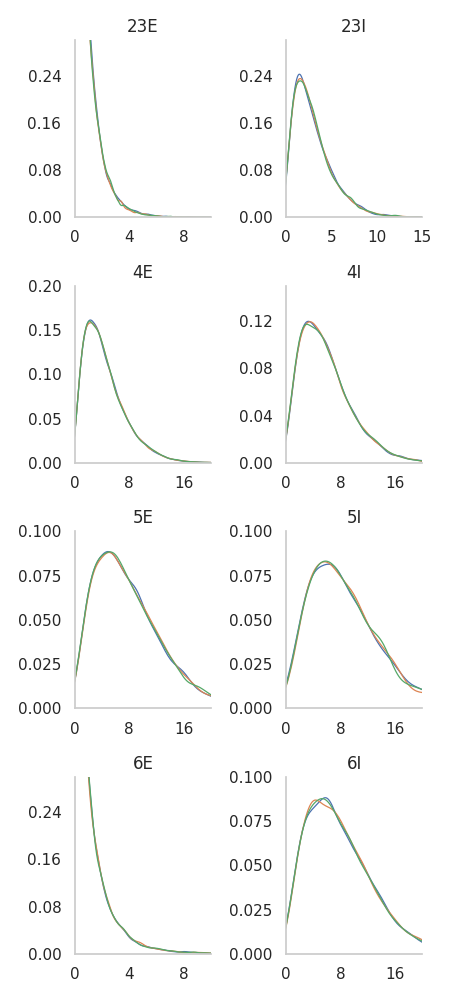}
    \caption{Spiking rate}
  \end{subfigure}
  \begin{subfigure}{.33\textwidth}
    \centering
    \includegraphics[width=0.9\linewidth]{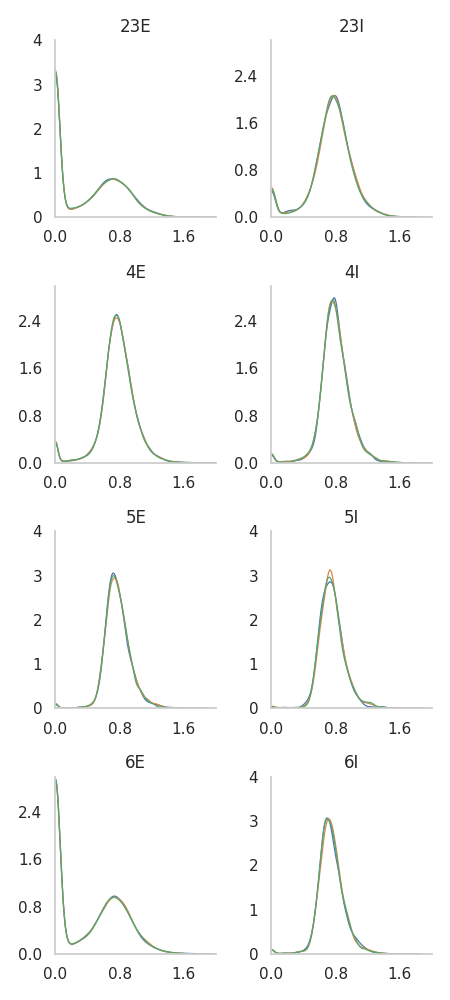}
    \caption{CV ISI}
  \end{subfigure}
  \begin{subfigure}{.33\textwidth}
    \centering
    \includegraphics[width=0.9\linewidth]{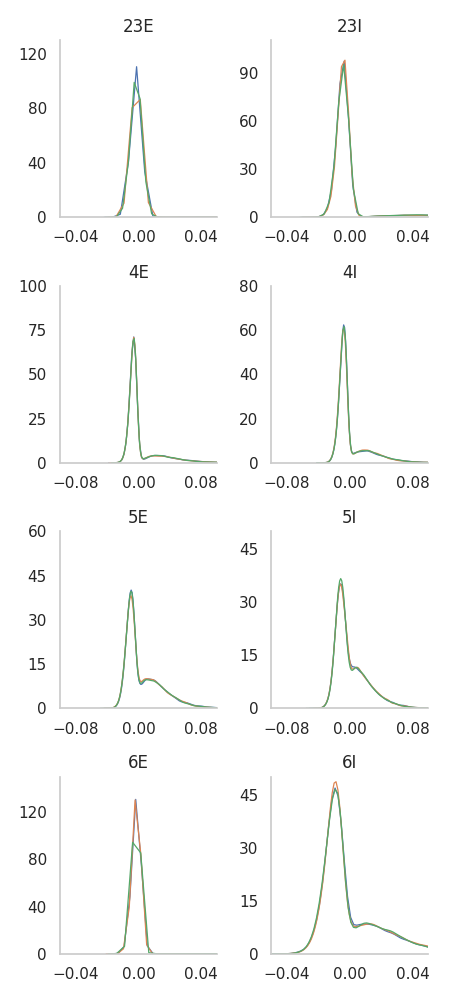}
    \caption{Pearson corr. coeff.}
  \end{subfigure}
  \caption{Kernel density estimates of \textbf{a)} spikes per second,
    \textbf{b)} covariance of interspike intervals, and \textbf{c)}
    Pearson correlation coefficient between binned spike trains for
    neuron samples. Blue lines for NEST, gold for gmem/mono/s, and
    green for jit/mono/s.}
  \label{fig:stats}
\end{figure*}
\begin{figure}
  \center
  \includegraphics[width=\linewidth]{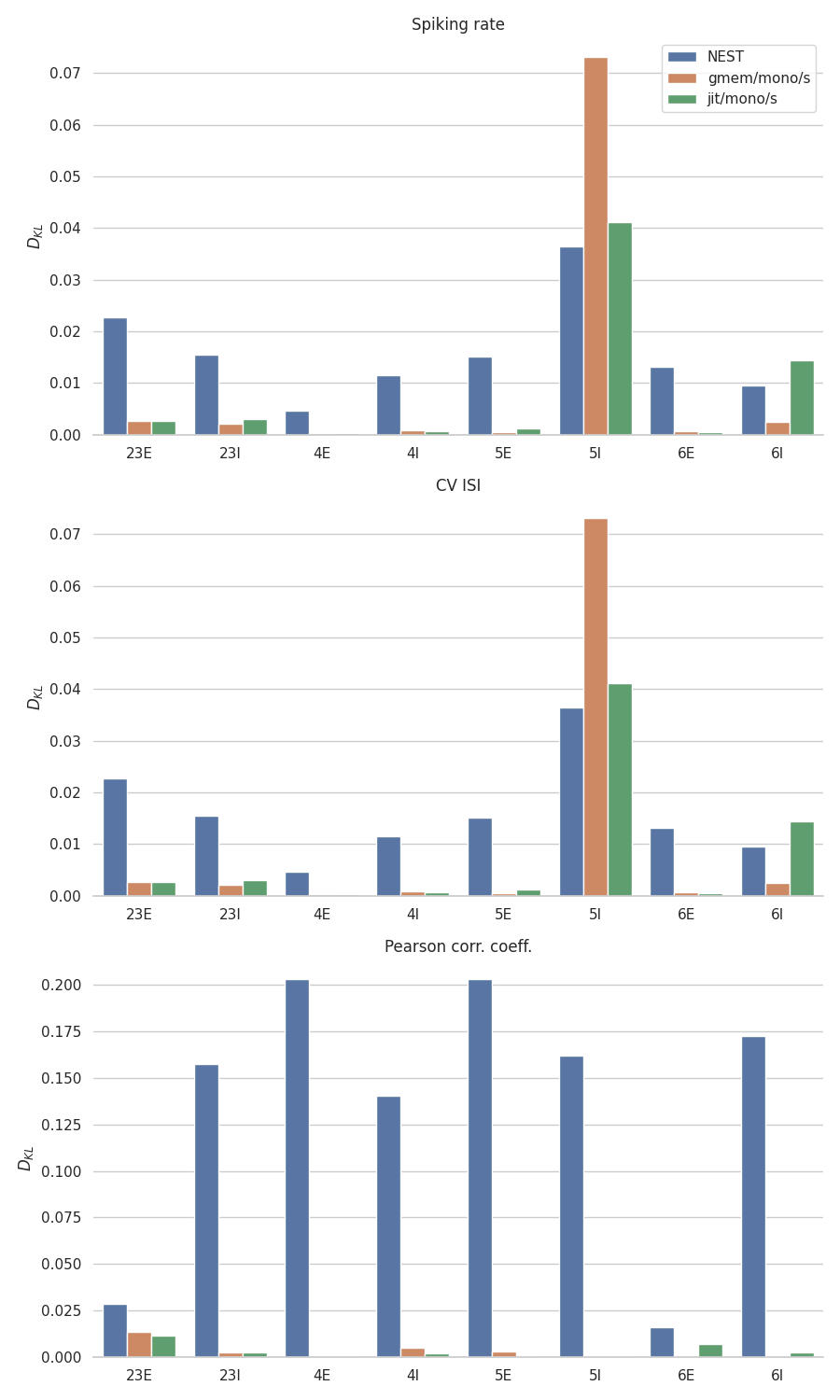}
  \caption{Kullback-Leibler (KL) divergence between distributions of
    single-neuron firing rates, CV ISIs, and Pearson correlation
    coefficients. Blue bars show KL divergences between NEST
    simulations run on different random seeds, gold and green bars
    between NEST and gmem/mono/s and jit/mono/s run on the same random
    seeds.}
  \label{fig:kldiv}
\end{figure}

We verify our simulators' correctness using the same methods
\cite{knight2018}; \cite{vanalbada2018}; \citep{potjans2014}; and
others used. I.e., we simulate the microcircuit initialized with the
same random seed with our simulators and with grid-based
(double-precision) NEST which we treat as our reference. We discard
spikes during the first second and compute the following statistics
over the remaining nine seconds over the eight populations; rate of
spikes per second, covariant of interspike intervals, and Pearson
correlation coefficient over binned spike trains. We smooth each
distribution with Gaussian kernel density estimation with bandwidth
selected using Scott's Rule.

\Cref{fig:stats} shows the distributions plotted for NEST in blue, the
gmem/mono/s family in gold, and the jit/multi/s family in green. The
plots indicate that the simulators produce distributions that are very
similar to NEST's which implies that the accuracy loss caused by the
reduced numerical precision is negligible. We do not plot the
distributions for our other families of single-precision simulators as
they are even closer to NEST's distributions. Neither do we plot
distributions for our double-precision simulators as they produce
results that are spike-for-spike identical to NEST. The
Kullback-Leibler (KL) divergence between the distributions, shown in
\cref{fig:kldiv}, quantifies the apparent similarities. The figure's
blue, gold, and green bars show the KL divergences between two NEST
simulations initialized with different random seeds, between NEST and
gmem/mono/s, and between NEST and jit/mono/s. The latter two
initialized with identical seeds. The divergence between two random
seeds are much larger than between NEST and our simulators, indicating
that they are accurate.

The non-associativity of IEEE 754 floating-point is the main reason
for the small differences. The order presynaptic current is added
depends on the spike transfer algorithm, which affects rounding. The
differences are stochastic and do not bias the result.

\subsection{Simulation speed}

\begin{table*}
  \centering
  \footnotesize
  \begin{tabular}{crrrrrrrrrrrr}
    \toprule
    \textbf{Simulator} & \textbf{UW} & \textbf{SU} & \textbf{H} & \textbf{SC} & \textbf{LC} & \textbf{RTF} & \textbf{Freq.} & \textbf{ALUT} & \textbf{Reg.} & \textbf{ALM} & \textbf{M20K} & \textbf{DSP}\\
    \midrule
    gmem/mono/s   &  8 &   2 & n/a & n/a & n/a & 4.79 & 601 & 126k & 313k & 21\% & 23\% &  1\%\\
    ''            & 64 &   2 & n/a & n/a & n/a & 4.93 & 585 & 155k & 391k & 26\% & 24\% &  2\%\\
    \specialrule{0.25pt}{1pt}{1pt}
    gmem/mono/d   &  8 &   1 & n/a & n/a & n/a & 5.38 & 608 & 141k & 371k & 24\% & 27\% &  2\%\\
    ''            &  8 &   2 & n/a & n/a & n/a & 4.71 & 608 & 147k & 394k & 25\% & 28\% &  2\%\\
    ''            & 16 &   2 & n/a & n/a & n/a & 4.61 & 605 & 182k & 477k & 30\% & 28\% &  4\%\\
    \specialrule{0.25pt}{1pt}{1pt}
    gmem/multi/s  &  4 &   4 & n/a & n/a & n/a & 4.91 & 600 & 122k & 294k & 20\% & 22\% &  0\%\\
    ''            & 64 &   1 & n/a & n/a & n/a & 5.65 & 585 & 128k & 340k & 22\% & 22\% &  6\%\\
    \specialrule{0.25pt}{1pt}{1pt}
    gmem/multi/d  &  8 &   1 & n/a & n/a & n/a & 5.52 & 609 & 139k & 365k & 23\% & 26\% &  2\%\\
    ''            &  8 &   2 & n/a & n/a & n/a & 4.86 & 605 & 145k & 360k & 24\% & 27\% &  2\%\\
    \specialrule{0.25pt}{1pt}{1pt}
    horiz/mono/s  & 32 & n/a &  16 &  32 & n/a & 0.81 & 608 & 127k & 329k & 22\% & 79\% &  4\%\\
    ''            & 64 & n/a &  16 &  16 & n/a & 0.85 & 604 & 143k & 375k & 25\% & 80\% &  7\%\\
    \specialrule{0.25pt}{1pt}{1pt}
    horiz/mono/d  &  4 & n/a &   8 &   4 & n/a & 1.74 & 609 & 122k & 333k & 21\% & 55\% &  1\%\\
    ''            & 32 & n/a &  16 &  32 & n/a & 0.82 & 601 & 258k & 653k & 42\% & 85\% &  9\%\\
    \specialrule{0.25pt}{1pt}{1pt}
    horiz/multi/s &  4 & n/a &   8 &   4 & n/a & 1.73 & 610 & 108k & 287k & 19\% & 51\% &  1\%\\
    ''            & 16 & n/a &  16 &  16 & n/a & 0.81 & 605 & 123k & 330k & 22\% & 80\% &  2\%\\
    ''            & 32 & n/a &  16 &  32 & n/a & 0.79 & 601 & 133k & 333k & 23\% & 81\% &  4\%\\
    \specialrule{0.25pt}{1pt}{1pt}
    horiz/multi/d &  4 & n/a &   8 &   4 & n/a & 1.79 & 583 & 126k & 319k & 22\% & 57\% &  1\%\\
    ''            & 16 & n/a &  16 &  16 & n/a & 0.80 & 607 & 187k & 492k & 32\% & 85\% &  5\%\\
    ''            & 32 & n/a &  16 &  32 & n/a & 0.79 & 600 & 265k & 664k & 43\% & 86\% &  9\%\\
    \specialrule{0.25pt}{1pt}{1pt}
    jit/mono/s    & 16 & n/a & n/a &  16 &  16 & 1.47 & 590 & 116k & 319k & 22\% & 79\% &  7\%\\
    \specialrule{0.25pt}{1pt}{1pt}
    jit/mono/d    &  4 & n/a & n/a &   4 &   8 & 2.23 & 611 & 122k & 317k & 21\% & 56\% &  2\%\\
    ''            & 16 & n/a & n/a &  16 &  16 & 1.44 & 597 & 182k & 478k & 32\% & 85\% & 10\%\\
    ''            & 32 & n/a & n/a &  32 &  16 & 1.50 & 584 & 268k & 704k & 44\% & 85\% & 20\%\\
    \specialrule{0.25pt}{1pt}{1pt}
    jit/multi/s   & 16 & n/a & n/a &  16 &  16 & 1.43 & 593 & 120k & 325k & 22\% & 79\% &  7\%\\
    \specialrule{0.25pt}{1pt}{1pt}
    jit/multi/d   &  4 & n/a & n/a &   4 &   8 & 2.21 & 600 & 125k & 317k & 22\% & 55\% &  2\%\\
    ''            &  8 & n/a & n/a &   8 &  16 & 1.56 & 602 & 147k & 393k & 26\% & 84\% &  5\%\\
    ''            & 16 & n/a & n/a &  16 &  16 & 1.34 & 606 & 186k & 492k & 33\% & 85\% & 10\%\\
    \bottomrule
  \end{tabular}
  \caption{Speed and resource usage for some simulators. The table
    includes each family's fastest simulator and some
    others for comparison purposes.}
  \label{tbl:perf3}
\end{table*}

\begin{figure}
  \center
  \includegraphics[width=\linewidth]{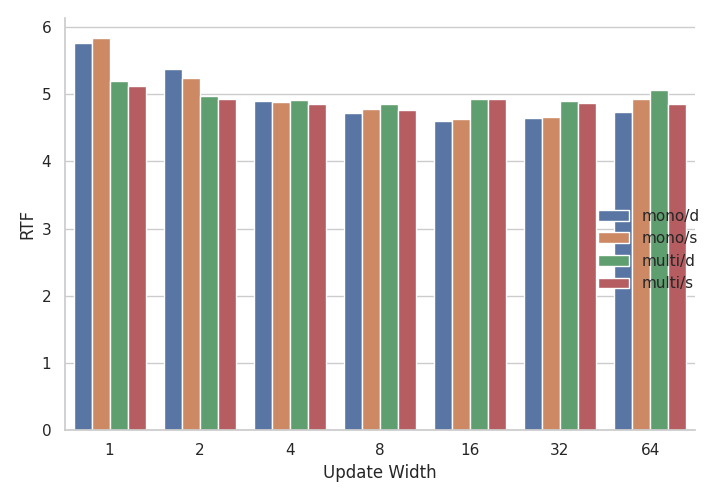}
  \caption{RTF as a function of update width for the fastest gmem/mono/s
    (blue), gmem/mono/d (gold), gmem/multi/s (green), and gmem/multi/d
    (red) simulators. Update widths larger than four does not decrease
    RTF.}
  \label{fig:uwidth}
\end{figure}

\begin{figure}
  \center
  \includegraphics[width=\linewidth]{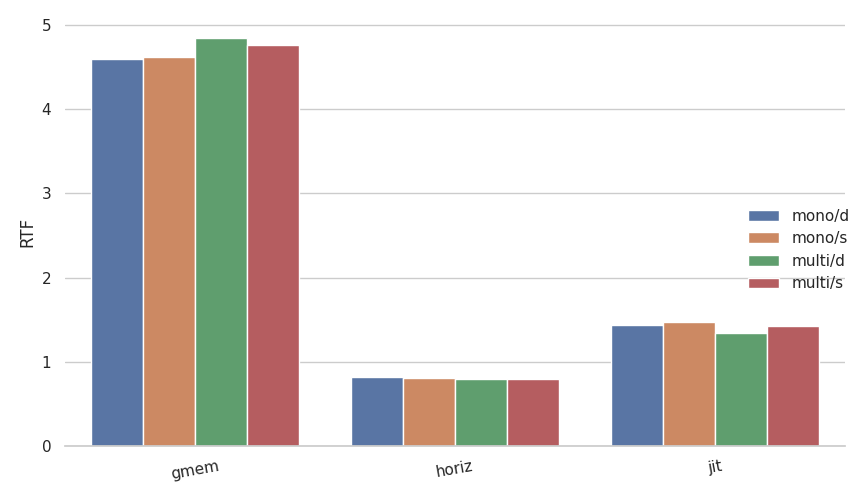}
  \caption{RTF of the fastest simulator of each type. The four colors
    represent the four implementation styles.}
  \label{fig:rtf}
\end{figure}

\Cref{tbl:perf3} presents the performance and resource usage of our
fastest simulator configuraions. The first column shows the
simulator's family in slash-notation (see \cref{sec:impl}). The next
five its parameters; \textit{update width} (UW), \textit{synapse
  unroll} (SU), \textit{horizon length} (H), \textit{synapse classes}
(SC), and \textit{lane count} (LC). The following column shows its
real-time factor (RTF), defined as the time taken to run the
simulation -- wall-clock time -- divided by the duration of the
simulated biological time (10 seconds). We measure the wall-clock time
as the time from the first \verb!clEnqueueNDRangeKernel! call until the
final simulation result can be read from the FPGA's memory. As the RTF
does not vary beyond two decimal places even on runs on networks
initialized with different random seeds, we just report its mean. The
next columns shows the simulator's operating frequency in MHz and FPGA
resource usage. \Cref{fig:rtf} plots the RTF of the fastest
simulators from every family as a bar chart.

A startling finding is that there is no significant performance
difference between double and single precision neuron state and
between multi- and single-kernel implementations. The fastest
horiz/mono/s and horiz/multi/d simulators RTF is 0.81 and 0.79 which
is very close to each other. Presumably, the spike transfer phase is
much more expensive than the neuron update phase so making the latter
run faster, either by using single-precision or by overlapping the
update and transfer phases, does not improve performance. For the same
reason, increasing the gmem simulator's \textit{update width} is
ineffective, as \cref{fig:uwidth} shows. Doubling the \textit{update
  width} roughly halves the number of cycles spent updating neuron
state, but even that is insignificant. The performance even detoriates
for \textit{update width} larger than 16. Likely because the
FPGA's DDR interface is 512 bits wide.

Neither the multi- nor single-kernel gmem simulators benefit from
more \textit{synapse unroll}. The reason could be because
the unrolled versions of the loop contains a false memory
dependency. There is no way of letting the OpenCL compiler know
that two iterations of the loop body -- \verb!W[t + d, j] += w!  --
writes to distinct memory locations so the compiler refuses to
schedule multiple writes per clock cycle.

The jit and horiz simulators are markedly faster than the gmem
simulators because they transfer spikes in on-chip memory. The best
gmem simulator has an RTF of 4.61 while it is 1.50 for the best jit
simulator and 0.79 for the best horiz simulator. The latter simulators
also uses up to 90\% of the on-chip memory for storing the horizon and
lane buffers. The results suggest that the larger these buffers are
the better the performance. Unfortunately, our FPGA can not fit
horizons longer than 16 time steps or more than 16 lanes. The horiz
simulators' performance edge over the jit simulators is due to them
running the spike transfer loop fewer times. The jit simulators run it
$d_\mathrm{max}$ times (i.e. 64) for every spiking neuron, while the
horizon simulators only runs it $d_\mathrm{max}/h$ times (i.e.,
64/16=4 times). The horiz simulators also does not have to sum the
incoming current from multiple lanes when updating the neurons state.

\begin{figure*}
  \center
  \includegraphics[width=\linewidth]{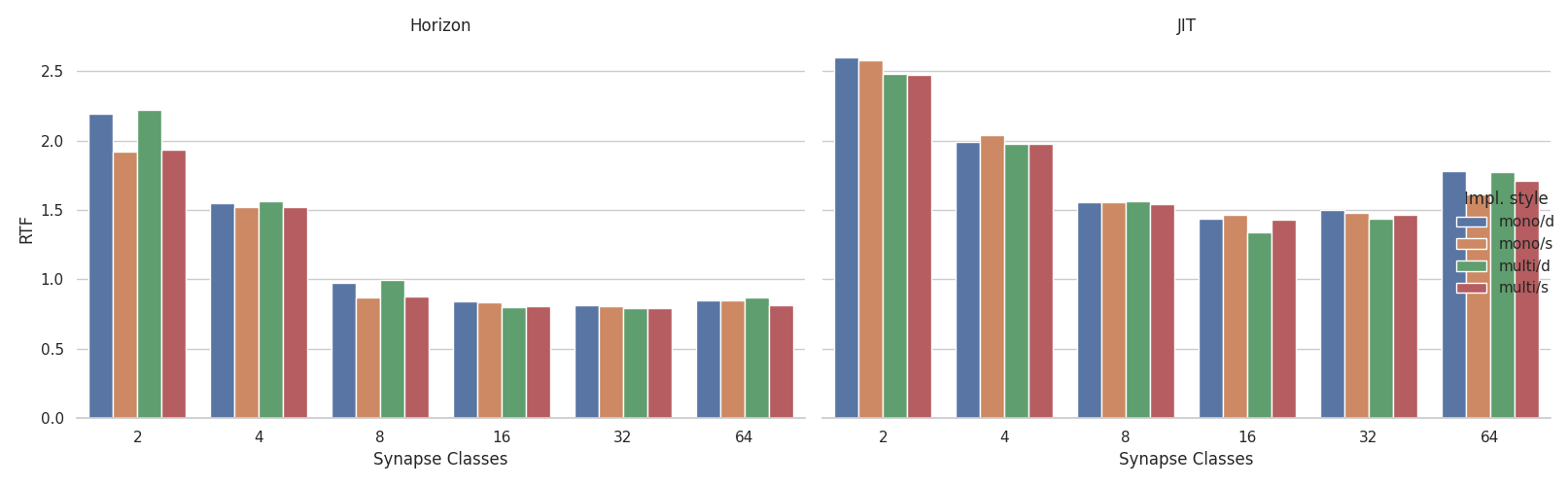}
  \caption{The horizon and JIT algorithms' RTF as a function of the
    number of \textit{synapse classes}. The \textit{horizon length}
    and \textit{lane count} parameters are both 16. The four colors
    represent the four implementation styles.}
  \label{fig:synapse_classes}
\end{figure*}

\Cref{fig:synapse_classes} plots RTF as a function of the number of
\textit{synapse classes} for the horiz and jit simulators. For both,
performance improves until the parameter reaches 32. The more classes,
the more synapses can be activated in parallel which, clearly, is
important for performance. The drawback of increasing the number of
classes is memory waste. Especially for the jit simulators which
iterate many fewer times per spike transfer loop. With 16 classes the
average occupancy is only about 74\%. It decreases to 47\% with 32
classes, meaning that the simulator accesses more than twice as much
data than it needs. Interestingly, there is no performance
penalty. The benefits of handling many synapses in parallel is worth
lots of extra off-chip memory reads. Perhaps since we store the synapses
in contiguous memory reading them is quite cheap.

All simulators run at around 600 MHz, considered very good for HLS. In
our experience, operating frequencies much lower than that were
caused by inefficiently banked on-chip memory, unaccounted for
loop-carried dependencies, or similar issues.

\subsection{Energy Usage}

We use the Terasic Dashboard GUI to measure the energy usage of our
fastest simulator -- horiz/multi/s with the parameters H=16, UW=32,
and SC=32. The dashboard connects to the Agilex 7 via a MAX 10 device
that continuously monitors the voltage and current rails going into
the board and the FPGA itself \citep{dkug}, allowing us to measure
both a ``pessimistic'' power consumption for the whole board (which
includes unused peripherals that draw power) and an ``optimistic'' one
for only the FPGA fabric. Due to the device's low sampling rate we
compute the total energy usage as the product of the maximum power
draw and the simulation time. According to our measurements, the
simulator pessimistically requires 44.9 W $\cdot$ 8.13 s = 101 mWh and
optimistically 16.3 W $\cdot$ 8.13 s = 37 mWh to simulate 10 seconds
of biological time. The pessimistic energy per synaptic event (metric
defined in \cite{vanalbada2018}) is 21 nJ and the optimistic one 9
nJ. These values are upper bounds and the actual energy usage may be
lower.

\section{Discussion}
\label{sec:disc}

Our main contribution is the presentation and analysis of methods for
creating FPGA-based simulators competitive in speed and energy usage
with the state-of-the-art in SNN simulation. We hope that others will
create even faster simulators by adopting and refining our algorithms
and implementation techniques for their hardware. Some methods are
endogenous to our hardware. For example, by running multiple kernels
in parallel that communicate with each other via channels, we overlap
different phases of the simulation algorithm. This technique has no
direct GPU-equivalent as different kernel types cannot communicate. It
also relies on the Intel-specific channel extension and may not work
well even on other vendors' FPGAs. On the other hand, the technique
for interleaving synapses to improve banking and reduce conflicts
(\cref{sec:disjoint}) is adaptable to non-FPGA hardware. The same goes
for the horizon-based transfer algorithm which should suit any device
with enough fast local memory.

Push-based SNN simulation mandates irregular memory accesses and has
low computational intensity; qualities that it shares with many other
graph processing problems. This is evidenced by the fact that most of
our simulators use less than 10\% of the FPGA's DSP resources, while
many consume over 80\% of its on-chip memory. More or even faster
arithmetic resources would be useless. However, more on-chip
memory would be very beneficial because we could use the basic spike
transfer algorithm and would not need to bother with the more complex
horizon algorithm. And higher bandwidth and lower latency memories
would allow the simulators to transmit spikes faster. This finding
implies that simulators would probably achieve excellent performance
with neuron models more complex than LIF -- as long as the
interactions between neurons remain the same and as long as their
memory consumption does not grow.

\subsection{FPGAs for HPC}

We believe that FPGAs have unique advantages for HPC; they can be
tailored for the problem at hand thanks to their
malleability. However, FPGAs and particularily \textit{tools for
  designing for FPGAs} have many glaring weaknesses. Unlike compilers
for software, which emit the same machine code for the same source
code, synthesizers use optimization algorithms based on randomness so
good performance is contingent on choosing lucky random seeds. Even if
the differences between lucky and unlucky seeds are small, the
randomness makes squeezing out the last few percent of performance
frustrating. An algorithmic change or parameter tuning causing a small
performance improvement may be due to chance. Moreover, a single
synthesis can take several hours even on top-of-the-line hardware,
compounding this problem. Writing performance-critical code is an
experimental process, wherein one needs to test hundreds or thousands
of ideas to see what yields the best performance. Long turn-around
times slows the process down. Simulators do not help as their
performance does not reflect the performance of real hardware. For
these reasons, FPGAs are decidedly more complex to develop for than
GPUs.

What is a good distribution of FPGA resources for HPC? For us, our
board's distribution is far from perfect. Our fastest designs used
almost all on-chip memory, with plenty of ALUTs, FFs, and DSPs to
spare. If our use-case and designs are close to the norm then it would
be wise for FPGA vendors to trade-off logic resources for more on-chip
memory. And trends in the HPC field indicate that memory -- not
arithmetic -- very much is a limiting factor for many
problems. However, inevitably, whatever resource distribution the
vendor chooses, it will not be ideal for some designs. Optimizing an
FPGA so that as many designs as possible can take advantage of as much
of its resources as possible seems exceedingly difficult.

\subsection{HLS for HPC}

In this work we choose HLS over traditional design methods because of
its purported productivity advantages. How much performance did HLS
cost us and how much longer would it have taken us to implement the
simulators in an HDL? The question is an instance of the classic
dichotomy between performance and productivity that appears in many
corners of computer science. The literature suggests that in general
the productivity gains of HLS are large and the performance losses are
either non-existent or low \citep{lahti2018, pelcat2016}. However, one
can ask whether this holds true for performance-critical design?

HLS definitely allowed us to explore algorithmic ideas at a rapid
pace. In particular, scheduling stallable loops and interfacing with
DDR would have been at least an order of magnitude more work to
implement (and debug!) in an HDL. It also helped that most -- but not
all -- of the OpenCL code we wrote were runnable verbatim on non-FPGA
targets. However, for performance judicious use of compiler
directivies is essential, something we struggled with. For example, adding
\verb!#pragma disable_loop_pipelining! on a loop in a gmem simulator
increased its operating frequency by almost 200 MHz since the compiler
could deduce that a memory port would not be shared across loop
iterations. On the other hand, removing the directive on a loop in a
horiz simulator more than doubled its performance! On several
occassions, misplaced \verb!#pragma ivdep! directives caused bugs that
were difficult to troubleshoot. Having to ``nudge'' the compiler via
directives to make the right decisions made us feel like we were not
fully in control and was at times frustrating.

As we have no baseline to compare with, estimating the performance
cost of HLS is difficult. Our designs run at over 600 MHz which --
while a fair bit lower than the theoretical limit -- is in the upper
range of what typical Agilex 7 HDL designs runs at. If we are correct
in that performance is mostly bounded by memory resources, then
non-optimal performance is due to algorithmic choices and not due to
the choice of implementation language.

\subsection{Future Research}

\begin{figure}
  \center
  \includegraphics[width=\linewidth]{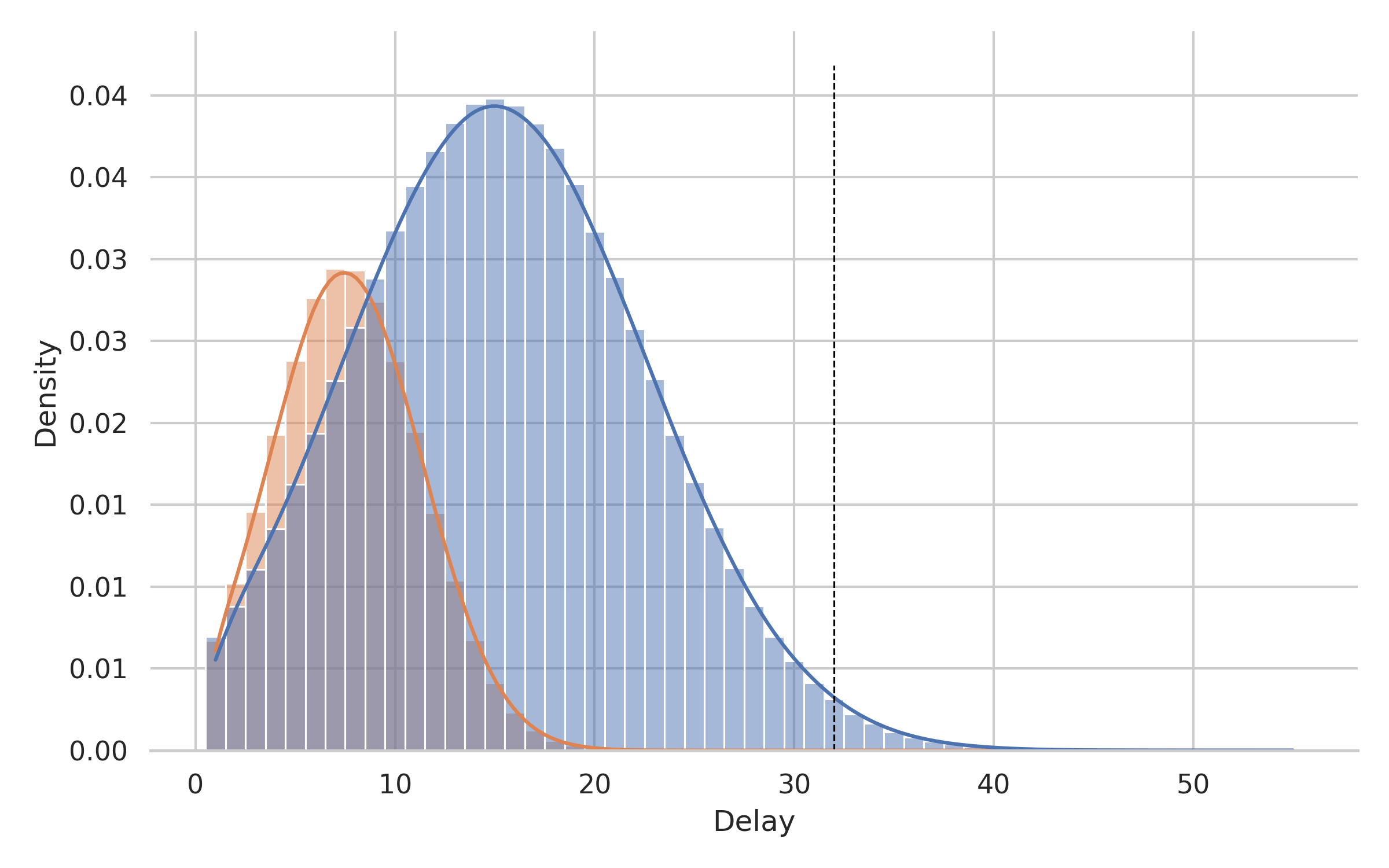}
  \caption{Synaptic delay distributions for excitatory (blue) and
    inhibitory synapses (gold). Almost all probability mass lies to
    the left of 32 (dashed line).}
  \label{fig:delays}
\end{figure}

Time constraints forced us to leave many ideas for better performance
unexplored. We list some of them here.

Our results demonstrate that single-precision floating-point is
sufficient. Half-precision or some 16-bit fixed-point representation
could also be sufficient. If so, four bytes would be enough to
represent synapses which would -- more or less -- cut our simulator's
memory demand in half.\footnote{Assuming we store some bits of the
delay and destination neuron index implicity in the index.}  And, as the
synapses current is sampled from Gaussians with known means, the
representation's size could perhaps be reduced further by storing the
current as \textit{deviations from the mean}. Presumably, fewer bits are
needed to represent the deviations accurately since they are rather
small.

Along the same lines, one could exploit the fact that, as
\cref{fig:delays} shows, synaptic delays are not uniformly
distributed. Something our algorithms are oblivious to. It means, for
example, that the gmem simulators needto use off-chip memory for the
spike transfer array. One could perhaps retain the spike transfer
array for ``fast'' synapses with delays shorter than, say, 32 and
another more space-efficient format for ``slow'' synapses with longer
delays.

As we have emphasized, we cannot safely transfer spikes from two
neurons simultaneoulsy as they may write to the same memory
locations. But this is only true if they have synapses with the same
destination neuron and delay. We can apply the idea from
\cref{sec:disjoint} on the ``neuron level'' and partition the
neurons into disjoint classes so that synapses of neurons of
different classes never trigger writes to the same memory locations.

Our FPGA has four DDR memories each equipped with a 512 bit read/write
port. We are likely not utilizing all bandwidth properly since we use
the compiler's default off-chip memory organization. Plausibly, we
could see a substantial performance improvement by carefully laying
out off-chip memory manually. For example, by distributing the
synaptic data (by far the largest data structure) over the four
memories.

Another line of research could be to deploy a cluster of FPGAs for SNN
simulation. As the communication between the devices likely becomes
the bottleneck, it is a much different problem from using one device.

\subsection{State-of-the-art}
\label{sec:sota}
\begin{table*}
  \centering
  \footnotesize
  \begin{tabular}{lllrrr}
    \toprule
    \textbf{Work} & \textbf{Simulator} & \textbf{Hardware} & \textbf{Node} & \textbf{RTF} & \textbf{Syn.Ev En.}\\
    \midrule
    \textbf{This work}       & \textbf{horiz/mono/s} & \textbf{1 Agilex 7 FPGA} &  \textbf{10} & \textbf{0.81} & \textbf{25}\\
    \cite{heittmann2022}     & IBM INC-3000          & 432 Xilinx XC Z7045 SoC &  28 &  0.25 &  783\\
    \cite{kauth2023b}        & neuroAIx              & 35 NetFPGA SUME         &  28 &  0.05 &   48\\
    \cite{golosio2021}       & NeuronGPU             & 1 GeForce RTX 2080 Ti   &  12 &  1.06 &  180\\
    \cite{golosio2021}       & NeuronGPU             & 1 Tesla V100            &  12 &  1.64 &    -\\
    \cite{knight2018}        & GeNN                  & 1 GeForce RTX 2080 Ti   &  12 &  1.40 &    -\\
    \cite{knight2018}        & GeNN                  & 1 Tesla V100            &  12 &  2.16 &  470\\
    \cite{vanalbada2018}     & SpiNNaker             & 217 ASIC                & 130 & 20.00 & 5900\\
    \cite{rhodes2020}        & SpiNNaker             & 318 ASIC                & 130 &  1.00 &  600\\
    \cite{kurth2022}         & NEST                  & 2 AMD EPYC Rome         &  14 &  0.53 &  480\\
    \bottomrule
  \end{tabular}
  \caption{State-of-the-art in hardware-accelerated microcircuit
    simulation. Node refers to the technology node in nanometers, RTF
    how much slower than realtime the simulator runs (lower is
    better), and Syn.Ev En. estimated energy consumption per synaptic
    event in nano-Joule.}
  \label{tab:power}
\end{table*}
\begin{figure}
  \center
  \includegraphics[width=\linewidth]{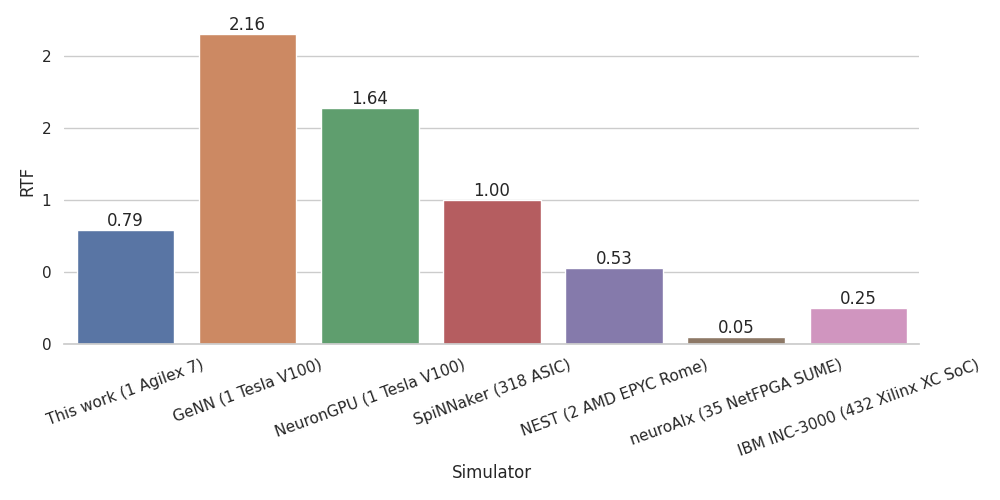}
  \caption{RTFs of some SNN simulators}
  \label{fig-rtf}
\end{figure}
\begin{figure}
  \center
  \includegraphics[width=\linewidth]{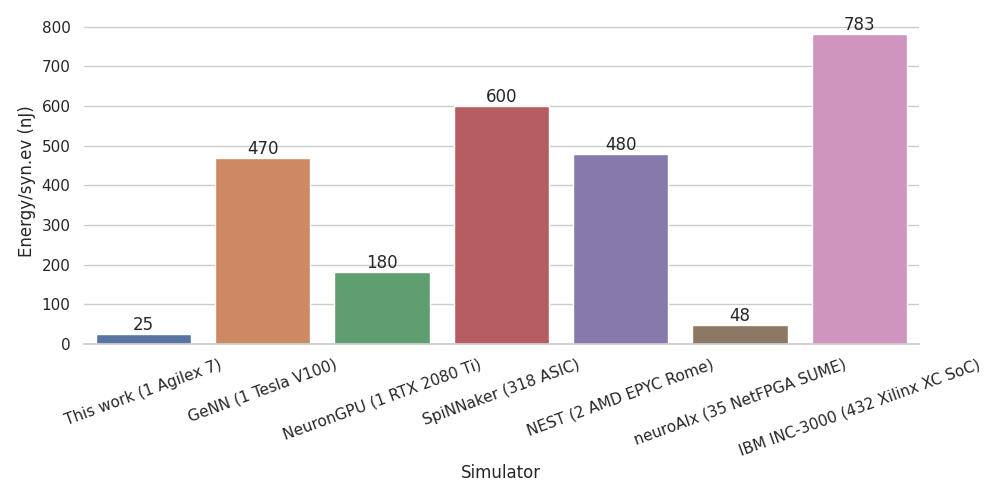}
  \caption{Energy per synaptic event of some SNN simulators}
  \label{fig-energy}
\end{figure}

While we are far from the first to have investigated FPGA-based SNN
simulation, our investigation is qualitatively different from most
others. First, our focus is accurate SNN simulation and not solving a
specific
task. \cite{gupta2020,han2020,li2021,zheng2021,carpegna2022,liu2023,carpegna2024}
all implement layered SNNs that excel at classifying images from the
MNIST dataset. For them classification accuracy is much more important
than simulation speed or energy consumption. Second, many FPGA
simulators simulate SNNs much smaller than the Potjans-Diesmann
microcircuit or uses non-LIF neurons, making their results
incomparable to ours. For example, \cite{pani2017} simulates up to 1,440
Izhikevich neurons in real-time and \cite{shama2020} simulates 150
Hodgkin-Huxley neurons on a Virtex-2 FPGA. In contrast, NeuroFlow by
\cite{cheung2016} simulates up to 600,000 neurons on a cluster of six
FPGAs four times slower than real-time. However, it uses a
1 ms timestemp (ten times larger than the \textit{de facto} standard
0.1 ms), organizes neurons into two-dimensional grids, and spatially
constrains synapses. That is, most synapses connect to the neuron's
nearest neighbours. \cite{trensch2022} proposes a simulator for a
Zynq-7000 SoC and measure its performance on a network of 800
excitatory and 200 inhibitory Izhikevich neurons. Their architecture
divides the FPGA into 16 identical processing blocks, each supporting
64 neurons or 1,024 neurons in total per chip. They argue that if a
cluster of their devices were deployed, it could simulate the
Potjans-Diesmann microcircuit seven times faster than
real-time. Though they did not test their claim.

The simulators we think are the most similar to ours are: IBM INC-3000,
neruoAIx, NeuronGPU, GeNN, SpiNNaker, and NEST. IBM-INC 3000 and
neuroAIx are both FPGA clusters, consisting of 432 Xilinx XC boards
and 35 NetFPGA SUME boards respectively. IBM-INC 3000 is four
times faster than real-time and neuroAIx twenty time faster than
real-time \citep{kauth2023,heittmann2022}. GeNN and NeuronGPU are both
GPU-based simulators written in CUDA and runs on NVIDIA's line of GPUs
\citep{knight2018,golosio2021}. On a Tesla V100 GPU, GeNN simulate
the microcircuit at about half the speed of real-time and NeuronGPU
5\% slower than real-time. SpiNNaker is an older dedicated
neuromorphic hardware system built out of hundreds of ASICs
\citep{furber2013}. NEST is a CPU-based simulator which runs
twice the speed of real-time on two AMD EPYC Rome nodes
\citep{kurth2022}.

\Cref{tab:power,fig-rtf,fig-energy} show how our best simulator stack
up against the competition. The energy usage of our fastest
simulator compares favorably to other results. Likely
partially due to to the Agilex 7's smaller technology node and to us
confining our implementation to one FPGA. \cite{kauth2023b} report a
much lower RTF than us, but use 35 boards, each drawing 26.54 W on
average, resulting in a higher total energy usage for the same amount
of biological time. They, like us, report an ``all-inclusive'' value
for the energy usage so a dedicated platform -- without any unused
peripherals -- could consume much less energy. It goes without
saying that fairly comparing performance of systems implemented on
different architectures, with different design trade-offs, and
accuracy constraints is tricky. One system with excellent performance
may be inadequate in other regards. For example, less configurability
may improve a system's performance, but make it unusable for certain
applications. It should be noted that most simulators, unlike ours,
replace thalamic spikes with DC input which limits their
applicability.

\section*{Author Contributions}
BAL and AP designed the study. BAL implemented the SNN framework and
performed the experiments. BAL and AP analyzed the results and
co-wrote the paper.

\section*{Funding}
This work was supported by the Swedish Research Council through the
project ``Building Digital Brains'' (grant reference 2021-04579).

\section*{Acknowledgments}
TODO

\section*{Supplemental Data}
 \href{http://home.frontiersin.org/about/author-guidelines#SupplementaryMaterial}{Supplementary Material} should be uploaded separately on submission, if there are Supplementary Figures, please include the caption in the same file as the figure. LaTeX Supplementary Material templates can be found in the Frontiers LaTeX folder.

\section*{Data Availability Statement}

All source code will eventually be available under a permissible Open
Source license.

\bibliographystyle{Frontiers-Harvard}
\bibliography{frontiers}

\end{document}